\documentclass[letterpaper]{article} 
\usepackage{aaai2026}  
\usepackage{times}  
\usepackage{helvet}  
\usepackage{courier}  
\usepackage[hyphens]{url}  
\usepackage{graphicx} 
\usepackage{natbib}  
\usepackage{caption} 
\usepackage{amsmath}
\usepackage{amssymb}
\usepackage{booktabs}
\usepackage{graphicx}
\usepackage{todonotes}
\usepackage{longtable}
\usepackage{microtype}
\usepackage{url}
\usepackage{booktabs}
\usepackage{todonotes}
\usepackage{placeins}
\usepackage{lineno}
\usepackage{xcolor}
\usepackage{appendix}

\usepackage{amsmath}
\usepackage{graphicx}
\usepackage[labelformat=simple]{subcaption}  
\usepackage{enumitem}  
\usepackage{pdflscape}  
\usepackage{multirow}  

\usepackage{algorithm}
\usepackage{algorithmic}

\usepackage{newfloat}
\usepackage{listings}

\floatstyle{ruled}

\floatname{listing}{Listing}

\title{Reducing the Scope of Language Models}

\author{
    David Yunis\thanks{David Yunis is a PhD student at the Toyota Technological Institute at Chicago. Work was performed during an internship at IBM.},
    Siyu Huo,
    Chulaka Gunasekara,
    Danish Contractor
}
\affiliations{
    IBM Research AI\\
    dyunis@ttic.edu,
    siyu.huo@ibm.com,
    chulaka.gunasekara@ibm.com,
    danish.contractor@ibm.com
}

\begin{document}

\maketitle

\begin{abstract}
Large language models (LLMs) are deployed in a wide variety of user-facing applications. Typically, these deployments have some specific purpose, like answering questions grounded on documentation or acting as coding assistants, but they require general language understanding. 
In such deployments, LLMs should respond only to queries that align with the intended purpose and reject all other requests, such as generating poetry or answering questions about physics, a task we refer to as `scoping'. We conduct a comprehensive empirical evaluation of various methods, ranging from prompting, fine-tuning to preference learning and the recently proposed general alignment technique known as Circuit Breakers (CB). Across three families of language models and a broad variety of tasks, we show that it is possible to scope language models. We examine scoping for multiple topics, and fine-grained topics. We ablate diversity of irrelevant queries, layer different techniques, conduct adversarial evaluations and more. Among other results, we find that when diverse examples of irrelevant queries are available, simple supervised fine-tuning produces the best results, but when such diversity is low, Circuit Breakers perform quite well. One can often get the benefits of both methods by layering them in succession. We intend our study to serve as a practitioner's guide to scoping LLMs. 
\end{abstract}
\begin{links}
    \link{Code}{https://github.com/IBM/llm-scoping}
\end{links}
\section{Introduction}

In recent years, large language models have surged into public awareness. One major recent addition is the ``alignment'' process through Reinforcement Learning with Human Feedback (RLHF)~\citep{christiano2017deep, ouyang2022training}, which has made the current generation of language models much less likely to emit toxic content than previous generations~\citep{wolf2017we}, and thus much more acceptable for general use. As a result, many businesses and individuals now feel more comfortable using these technologies than they did in the past.

Although we have generally capable language models that can refuse to answer toxic or dangerous queries, deploying them still remains challenging. Even if they avoid producing harmful content, they often respond to any question, relevant or not, without discernment. This becomes a problem when we wish to use them in specific contexts: e.g. shopping bots currently give coding advice\footnote{\url{https://shorturl.at/qf3FA}} or answer other questions,\footnote{\url{https://www.forbes.com/sites/lesliekatz/2024/07/13/amazon-ai-shopping-assistant-rufus-answers-non-shopping-questions-too/}} while assistive copilots can be taken off course by prompt injections.\footnote{\url{https://oecd.ai/en/incidents/2025-10-08-0fbf}} Thus, there is still a need to scope language models for these specific uses.

\begin{figure*}[!t]
  \centering
    \includegraphics[width=0.9\linewidth]{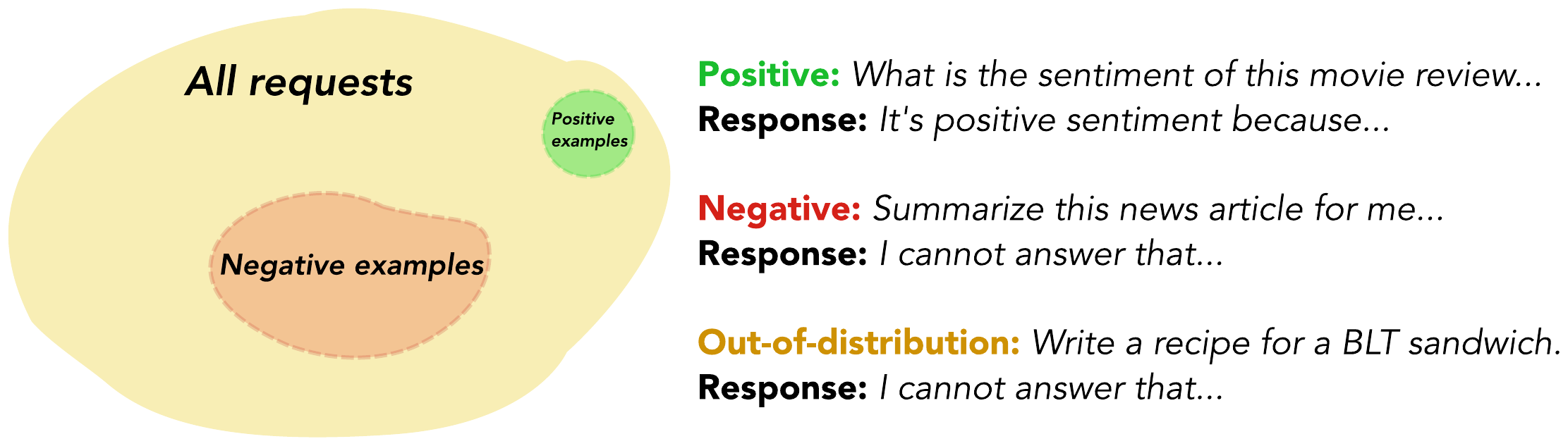}
  \caption{We study the ability to scope language models to specific topics. We assume access to a set of relevant (accept) queries and irrelevant (reject) queries, where the accept queries correspond to a relatively narrow domain. We examine how well different methods cause the language model to accept only the relevant examples, while rejecting all other examples, including out-of-distribution requests that weren't seen during training.}
  \label{fig:teaser}
\end{figure*}


We define LLM \textit{scoping} as a conditional generation task in which a language model must: (i) identify whether an input query falls within a relevant domain, (ii) reject irrelevant queries, and (iii) maintain high-quality generation for relevant queries. This contrasts with traditional text classification, which maps inputs to discrete labels without requiring natural language generation.


Let $\mathcal{Q}$ denote the set of all possible natural language queries. Let $\mathcal{D}_{\text{rel}} \subset \mathcal{Q}$ denote the subset of \emph{relevant} or \emph{in-domain} (should be \emph{accepted}) queries, and let $\mathcal{D}_{\text{irr}} = \mathcal{Q} \setminus \mathcal{D}_{\text{rel}}$ denote the set of \emph{irrelevant} or \emph{out-of-domain} (should be \emph{rejected})  queries. Let $\mathcal{R}$ be the space of valid natural language responses, and let $\bot \notin \mathcal{R}$ be a special token representing rejection.

The scoping is a function:
\[
f_{\theta}: \mathcal{Q} \rightarrow \mathcal{R} \cup \{\bot\}
\]
such that:
\begin{align*}
    f_{\theta}(q) &\in \mathcal{R}, \quad \text{if } q \in \mathcal{D}_{\text{rel}}, \\
    f_{\theta}(q) &= \bot, \quad \text{if } q \in \mathcal{D}_{\text{irr}}.
\end{align*}

In contrast, text classification models are not required to preserve generation quality, whereas in LLM scoping, degrading the model's performance on $\mathcal{D}_{\text{rel}}$ constitutes a failure of the task.
Further, unlike the traditional LLM refusal task in the context of safety and alignment, which typically corresponds to a much smaller reject set than accept set, the scoping tasks consider the opposite.

Currently, LLMs can be scoped through two-stage approaches like relevance classifiers, or system prompting and then text generation, but these options are brittle~\citep{chao2023jailbreaking, mehrotra2023tree, zeng2024johnny, wei2024jailbroken} and easy to circumvent. We shed further light on this problem, conducting a comprehensive empirical study on scoping language models to specific tasks. We apply existing methods to this problem, including system prompting, supervised fine-tuning, preference learning~\citep{rafailov2024direct}, probing, and a recently-introduced method called Circuit Breakers~\citep{zou2024improving}. We scope language models more broadly for multiple tasks, and more finely for specific niche tasks. We ablate over diversity of training sets, language model size, adversarial prompting and more. Our specific contributions include:

\begin{itemize} 
\item We introduce the task of scoping LLMs
\item We conduct a broad experimental exploration of existing methods for this task
\item We show that it is possible to scope language models, even for multiple and fine-grained tasks
\item We find that when training data exhibits high diversity, supervised fine-tuning yields the best performance
\item Conversely, in settings with low data diversity, the Circuit Breakers method~\citep{zou2024improving} provides superior results
\item Finally, we show it is possible to layer these two, often preserving the best performance
\end{itemize}

\section{Related Work}

\begin{table*}[ht]
\centering
\small
{

\begin{tabular}{|l|l|r|r|r|}
\toprule
Category & Example task & \# Datasets & \# Tasks & \# Instances \\
\midrule
Sentiment Analysis (SA) & Predicing if a movie review is relevant or not & 8 & 10 & 31248 \\
Toxic Language Detection (TLD) & Detecting if a comment contains cursing & 5 & 9 & 33849 \\
Summarization (S) & Condensing a news article & 4 & 4 & 13096 \\
Text Completion (TC) & Filling in the blanks in a transcript & 3 & 3 & 10515 \\
Story Composition (SC) & Writing a new ending for a story & 4 & 4 & 15556 \\
Dialogue Generation (DG) & Continuing a dialogue between parties & 3 & 4 & 12744 \\
Program Execution (PE) & Computing the result of a described function & 26 & 26 & 94001 \\
\midrule
Question Answering (QA) & Answering biology multiple-choice questions & 19 & 30 & 84065 \\
GSM8k~\citep{cobbe2021training} & Answering simple math word problems & 1 & 1 & 5978 \\
Alpaca~\citep{taori2023alpaca} & General requests like writing a recipe for lunch & 1 & - & 18793 \\
\bottomrule
\end{tabular}
}
\caption{Breakdown of data. We reserve at least 20\% of the data from each dataset for validation. We will use at most 2048 instances from each category for training, though this is sampled from a much larger number. PE is so large as the data is synthetically generated. All categories above the divider will be used for training and evaluation, while categories below the divider are only used for out of distribution evaluation.}\label{tab:data}
\end{table*}

\textbf{Aligning Language Models:} The advent of the current era of language models has been marked by a process of aligning language models so that generations are more helpful, and safer for deployment~\citep{ouyang2022training, bai2022training}. The primary way this is accomplished is through reinforcement learning with human feedback (RLHF)~\citep{christiano2017deep} which was first proposed in robotic simulation tasks. RLHF proceeds by collecting preference pairs of completions, and training a reward model from human judgments on those preference pairs, then performing reinforcement learning with the language model against that reward model. From tasks in simulation, it was developed in language~\citep{stiennon2020learning}, until it reached its current state. Other works have removed the human aspect of human feedback, allowing for synthetic feedback from models~\citep{bai2022constitutional, sudalairaj2024lab}. Lately, \citet{rafailov2024direct} have removed the need for a reward model, making for a stabler and simpler objective function without many of the complexities of RL training. A budding line of work also explores aligning not just to a single reward model, but preferences of many different individual users~\citep{chakraborty2024maxmin, lee2024aligning}. All of these methods focus on some general notion of alignment, without considering the specific task, unlike our work.

\textbf{Adapting for Specific Purposes:} Typically after pretraining, language models go through an instruction fine-tuning stage, where they gain the ability to follow instructions~\citep{mishra2022cross, ouyang2022training, wei2022finetuned}. After this, they proceed through an alignment phase as discussed above, usually to avoid harmful behavior~\citep{bai2022training}. It is possible to adapt language models for specific purposes simply with a system message~\citep{touvron2023llama}, but many examples of black-box adversarial attacks~\citep{chao2023jailbreaking, anil2024many, wei2024jailbroken, zeng2024johnny} demonstrate it is difficult only to rely on the system prompt for such control. \citet{wallace2024instruction} propose finetuning with different levels of priority, similar to \citet{zhang2023defending}, but these works focus primarily on general safety and not the task. These ideas are based on the fact that current language models can often be distracted by irrelevant context~\citep{shi2023large, yoran2024making}. Thus, it seems important to finetune the language model if we want it to be deployed to a particular domain. For domains where there is sufficient data, we may also pretrain and fix the language model's purpose ahead of time~\citep{beltagy2019scibert, wu2023bloomberggpt, li2023starcoder} or continue pretraining from a base language model~\citep{gururangan2020don}. It is an open question however whether finetuning retains the robustness capabilities, or if it is similarly as brittle as system prompting for out-of-distribution questions.

\textbf{Refusal in Language Models:} As our work deals with scoping models to refuse irrelevant queries, we review refusal. More detail is available in a comprehensive survey by \citet{wen2024art}. One common case to train for refusal is when the answer is unknown or the model is unconfident~\citep{zhang2023r, cao2023learn, xu2024rejection}. Another is for unsafe inputs~\citep{varshney2023art, zhang2023defending, wallace2024instruction}. Supervised fine-tuning (SFT) to reject unsafe prompts can still lead to unsafe behavior, though parameter efficient methods like LoRA~\citep{hu2022lora} have better tradeoffs~\citep{brahman2024art}. Both \citet{brahman2024art} and \citet{cheng2024can} take an approach to refusal using SFT and DPO, which we will adapt to our case. Other methods to induce refusal may be prompt-based~\citep{xie2023defending, zhang2024intention} or based on probing model representations~\citep{kadavath2022language, slobodkin2023curious}. \citet{zou2024improving} design a method that conditionally rejects unsafe inputs based on orthogonalizing internal representations that we will adapt for our study. Though these methods lay out a set of techniques to explore for our task, all of them are oriented toward general alignment qualities like safety, as opposed to specific tasks that we will explore.

\section{Experimental Setup}\label{sec:setup}

 \textcolor{black}{
 We would like to scope language models to provide completions to relevant tasks, and reject queries corresponding to irrelevant tasks. In particular, we assume we are given a set of ``relevant'' or ``accept'' queries $\{q_{rel} | q_{rel} \sim \mathcal{D}_{rel}\}$, where $\{\mathcal{D}_{rel}\}$ is a set of {\em accepted} tasks, and a set of ``irrelevant'' queries  $\{q_{irr} | q_{irr} \sim \mathcal{D}_{irr}\}$ where $\{\mathcal{D}_{irr}\}$ is a set of {\em rejected} tasks. We are given a language model $f_\theta : q \mapsto y$ which predicts completion $y$ from input $q$, with parameters $\theta$; a classifier $g: y \mapsto c \in \{0, 1\}$ which decides whether a LLM completion is accepted (0) or rejected (1). We would like to compute an update $\Delta$ such that we minimize $\mathop{\mathbb{E}}_{q_{rel}} g(f_{\theta + \Delta}(q_{rel}))$ and maximize $\mathop{\mathbb{E}}_{q_{irr}} g(f_{\theta + \Delta}(q_{irr}))$. Thus we want `relevant' queries to be accepted and `irrelevant' queries to be rejected.}

\textcolor{black}{
As an additional goal of scoping, we would like performance on the accept tasks not to degrade. Given a scoring function $h: (q, y) \mapsto s \in [0, 1]$ which scores the completion on task performance where 1 is best, we would also like to maximize $\mathop{\mathbb{E}}_{q_{rel}} h(q_{rel}, f_{\theta + \Delta}(q_{rel}))$.}

\subsection{Datasets \& Metrics}

We conduct many experiments with different mixtures of accept and reject queries. In order to standardize the format, we draw prompts from Super-NaturalInstructions (SNI)~\citep{wang2022super}. SNI is a meta-dataset composed of many different ``tasks'', sometimes with multiple tasks per dataset, for example generating questions from passages for a reading comprehension dataset, or generating answers to provided questions from the same reading comprehension dataset. Each task, specified by a task instruction, comes with a collection of examples. We use SNI as it is publicly available, and contains a broad range of complex tasks which current language models should be able to perform. To get our training datasets, we first manually select a set of tasks that are straightforward to automatically evaluate, leaving out many more subjective tasks that may require a human reader. We then group those tasks that we select by category provided from SNI. Details and statistics on categories are provided in Table~\ref{tab:data}.

Each of these categories contains multiple datasets, so the distribution for each task is quite broad. We will also combine multiple tasks in the accept or reject set. For all experiments, we always evenly split the training data for accept/reject set between all tasks. We reserve at least 20\% of the prompts as a validation set that are not seen during training. Where not specified, we use 2048 prompts for the accept set, and 2048 prompts for the reject set. We evaluate Sentiment Analysis and Toxic Language Detection with accuracy (the classes are mostly balanced), while for other tasks we use a standard metric for generation, Rouge-L~\citep{lin2004rouge}, between the generation and ground truth completion as a proxy for performance (Accept Score). Our goal is mostly to study rejection behavior, so a rough performance proxy is all we need.

Irrespective of the specific accept and reject sets used during training, we evaluate performance across all categories listed in Table~\ref{tab:data}. We then report the average rejection rates separately for the accept set (Accept), all in-distribution reject sets (ID Reject), and all out-of-distribution reject sets (OOD Reject). Note if the training set consists of SA in Accept, and S in Reject, OOD Reject will contain the 8 other categories. Tasks above the divider in Table~\ref{tab:data} will be used in different experiments for both training and evaluation, while tasks below the divider will only be used for OOD evaluation. As stated previously, ideally we would like to have 0 rejection on Accept, and 100\% rejection on ID Reject and OOD Reject.

\subsection{Methods}

For all methods that require training the language model, we use LoRA~\citep{hu2022lora} training with rank $16$, $\alpha = 16$ and dropout of $0.05$. We use the Adam~\citep{kingma2014adam} optimizer without any regularization and tune learning rates (see Appendix B).

\textbf{System Prompting (Sys.):} The simplest method to scope language models is simply to instruct them to refuse irrelevant prompts. For example, for SA the system prompt is: \textit{You are an assistant who only answers requests related to Sentiment Analysis. For all other requests you respond ``I cannot answer that.''} With multiple accept categories, we comma separate the category names (e.g. \textit{``...related to Sentiment Analysis, Text Completion and Summarization...''}). This system prompt is prepended to all instructions at evaluation time. In addition, all other methods also use the system prompt both at training and evaluation time. This is similar to methods proposed by \citet{xie2023defending, zhang2024intention}.

\textbf{Supervised Fine-Tuning (SFT):} Supervised Fine-Tuning (SFT) consists of tuning the language model to produce particular outputs. For the accept tasks the completions $y_{rel}$ are the groundtruth completions provided by the dataset. For the reject tasks, the completions $y_{irr}$ are always \textit{``I cannot answer that."}. We tune learning rate and step budget for SFT. This is a similar approach to \citet{brahman2024art, cheng2024can}. In experiments, the $\mathcal{L}_{gen}$ is about generation (task completion) loss between $y_{rel}$ and $f_\theta(q),{q \in \mathcal{D}_{\text{rel}}}$, and the $\mathcal{L}_{rej}$ is about irrelevant task rejection loss between $y_{irr}$ and $f_\theta(q),{q \in \mathcal{D}_{\text{irr}}}$. The total scoping loss is $\mathcal{L}_{scope} =  \mathcal{L}_{gen} + \lambda \mathcal{L}_{rej}$.  We use balanced scoping loss where $\lambda=1$ for loss computing.  As the finetuning dataset can be quite small, loss is only computed on the completions so as to avoid overfitting to the small set of instructions, agreeing with common practice~\citep{mishra2022cross, ouyang2022training, wei2022finetuned}. 

\textbf{Direct Preference Optimization (DPO):} Given its role in post-training we explore a preference learning method. We choose to experiment on Direct Preference Optimization (DPO)~\citep{rafailov2024direct}, as it does not require an additional reward model. DPO requires pairs of preference data, so for accept queries we provide the dataset completion as preferred, and the completion \textit{``I cannot answer that."} as rejected. For reject queries we do the reverse, preferring \textit{``I cannot answer that."} over the ground truth completion. For DPO we tune learning rate, step budget, and the loss weighting term regularizing the KL divergence from the base model predictions. This is similar to \citet{brahman2024art, cheng2024can}.

\textbf{Two-stage on Probing Classifier (Probe):} Probes of representations are a common method to accomplish tasks as they base predictions on the internal state of the language model~\citep{conneau2018cram, tenney2019you, zou2023representation}. Previous work on Circuit Breakers~\citep{zou2024improving} showed that probing representations was competitive for detecting dangerous language. However, they only designed probes on a single layer of a language model. Here we design a stronger probe. Once an instruction is fed to the  frozen language model, we first remove the first position as it is quite anomalous due to large magnitude~\citep{xiao2023efficient}, then we average all positions per layer and normalize the average vector to norm 1 so as to match norms between layers. Finally, we concatenate the mean-pooled representations from each layer to form a single feature vector, which is then passed through a two-layer multilayer perceptron (MLP) with a hidden dimension of 256 to perform binary classification, determining whether to accept or reject the input. Only the MLP layers are trained, and we tune the learning rate and step budget. This is justified by prior choices in \citet{kadavath2022language, slobodkin2023curious, zou2024improving}. We use this probe in a two-stage setup where the probe first classifies the request and if it is determined to be an `accept' task,  a response generated via an LLM.

\textbf{Circuit Breakers (CB):} \cite{zou2024improving} first introduce a method they call Circuit Breakers (CB) for accepting normal queries while rejecting dangerous ones. We repurpose their method for this task. Essentially, given a function which extracts the representations of a language model at particular layers, they design an optimization objective with two components: $\mathcal{L}_{gen,{q \in \mathcal{D}_{\text{rel}}}}(q, \Delta) = \lVert \text{rep}(f_{\theta}(q)) - \text{rep}(f_{\theta+\Delta}(q)) \rVert_2^2 $ and $\mathcal{L}_{rej,{q \in \mathcal{D}_{\text{irr}}}}(q, \Delta) = \max \{0, \cos(\text{rep}(f_{\theta}(q)), \text{rep}(f_{\theta+\Delta}(q))\}$. The total loss is $\mathcal{L}_{scope} = \alpha(t) \mathcal{L}_{gen} + \beta(t) \mathcal{L}_{rej}$ where the two components of the loss are scheduled over time.

This loss function keeps the representations of accept tasks from drifting, while making the representations of reject tasks orthogonal from their original position. This orthogonalization breaks the language model generation on bad inputs. For CB we tune learning rate and step budget and more, and show later that CB is particularly sensitive to hyperparameters.

\textbf{SFT $\rightarrow$ CB:} As we will see, SFT and CB tend to be the best methods for scoping in slightly different circumstances. In order to improve accept task performance and preserve the benefits of both, we propose to layer CB on top of SFT. We first run SFT, then run CB training afterwards. We keep hyperparameters from the SFT and CB tuning respectively.

\subsection{Detecting rejection}
\begin{figure*}[!t]
  \centering
  \includegraphics[scale=0.45]
  {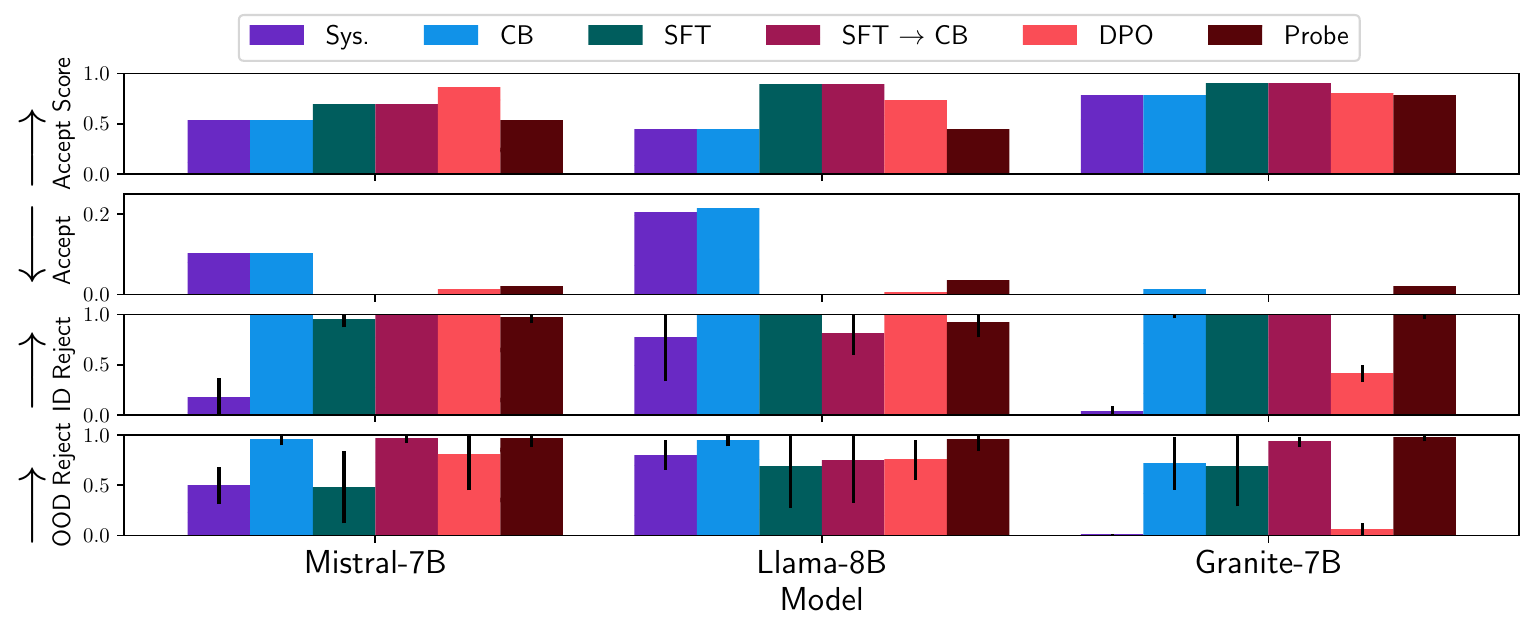}
  \caption{Scoping across different language models. We see that system prompting is insufficient, and different methods have different success rates for different models. Clearly it is possible to scope language models to particular distributions.}
  \label{fig:models}
\end{figure*}

\begin{figure*}[!t]
  \centering
  \begin{subfigure}[b]{0.3333\linewidth}
    \centering
    \includegraphics[width=\linewidth]{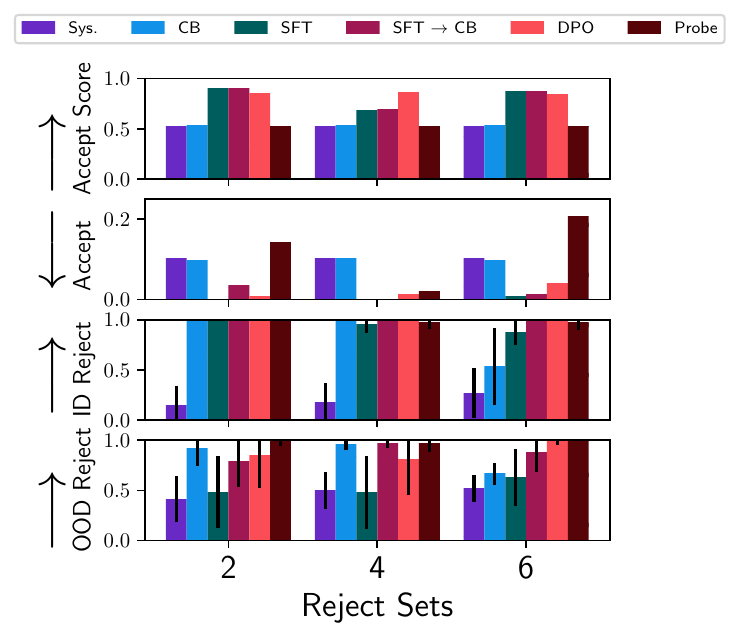}
    \caption{Mistral-7B-Instruct}
  \end{subfigure}\hfill\begin{subfigure}[b]{0.3333\linewidth}
    \centering
    \includegraphics[width=\linewidth]{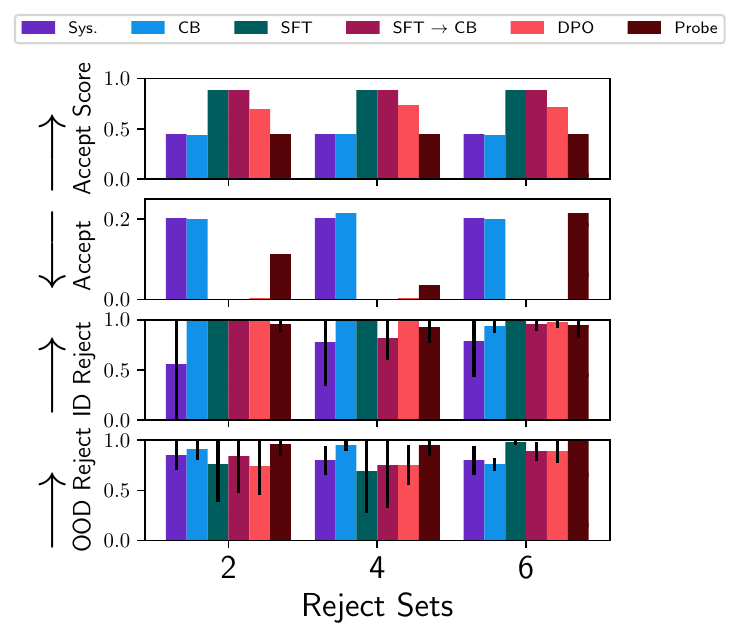}
    \caption{Llama-3.1-8B-instruct}
  \end{subfigure}\hfill\begin{subfigure}[b]{0.3333\linewidth}
    \centering
    \includegraphics[width=\linewidth]{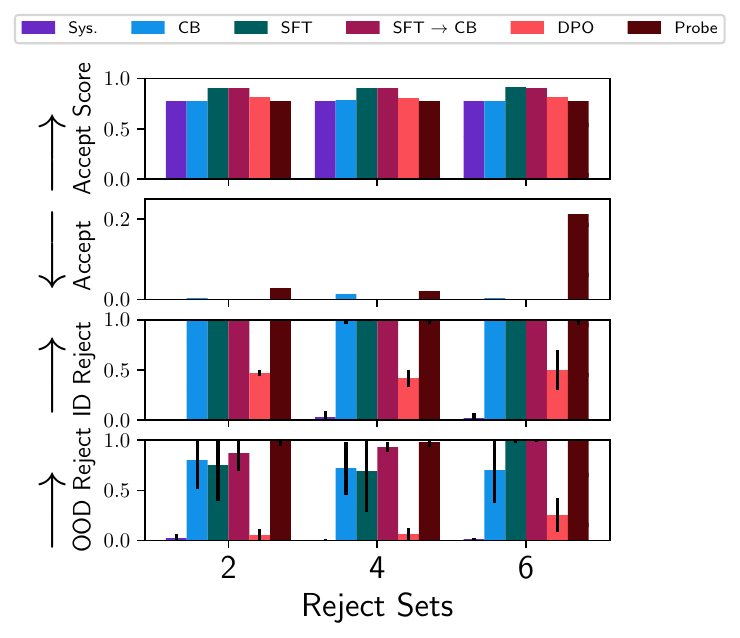}
    \caption{Granite-7B-instruct}
  \end{subfigure}
  \caption{Results for increasing diversity of rejection set. We see across models that CB performs relatively better than SFT when data diversity is low, but SFT is much stronger with more rejections sets. Probing appears strong across the board, though sometimes leads to overrejection on the Accept set.}
  \label{fig:diverse}
\end{figure*}

Though a strong language model judge~\citep{zheng2023judging} may seem like a good choice for judging rejection, we followed prior work in first experimenting with simpler heuristics~\citep{zou2023universal, zou2024improving} based on string matching and the rejection behavior of Circuit Breakers. We found these heuristics to be very strong, and much less expensive than running a frontier judge for the many experiments in this paper. We also experimented with a smaller hosted language model judge, but found its performance much poorer than the heuristics. More details on evaluation are available in Appendix B.
\begin{table*}[ht]
  \centering
  \resizebox{0.9\textwidth}{!}{
  \small
  {
    \begin{tabular}{l rrrr rrrr}
        \toprule
        & \multicolumn{4}{c}{Classification and Generation} & \multicolumn{4}{c}{Math and Programming} \\
        \cmidrule(lr){2-5} \cmidrule(lr){6-9}
        Method & Accept Score & Accept & ID Reject & OOD Reject & Accept Score & Accept & ID Reject & OOD Reject \\
        \midrule
        Sys. & $0.25 \pm 0.18$ & $0.10 \pm 0.11$ & $0.70 \pm 0.04$ & $0.42 \pm 0.11$ & $0.15 \pm 0.14$ & $0.16 \pm 0.22$ & $0.33 \pm 0.36$ & $0.28 \pm 0.22$ \\
        \midrule
        CB & $0.25 \pm 0.18$ & $0.10 \pm 0.12$ & \textbf{1.0 $\pm$ 0.0} & $0.79 \pm 0.23$ & $0.14 \pm 0.15$ & $0.16 \pm 0.22$ & \textbf{1.0 $\pm$ 0.0} & \textbf{0.96 $\pm$ 0.08} \\
        \midrule
        SFT & \textbf{0.46 $\pm$ 0.24} & \textbf{0.01 $\pm$ 0.02} & $0.95 \pm 0.03$ & $0.28 \pm 0.16$ & $0.26 \pm 0.04$ & \textbf{0.0 $\pm$ 0.0} & $0.99 \pm 0.01$ & $0.52 \pm 0.25$ \\
        \midrule
        SFT $\rightarrow$ CB & \textbf{0.46 $\pm$ 0.24} & $0.07 \pm 0.11$ & \textbf{1.0 $\pm$ 0.0} & $0.54 \pm 0.22$ & \textbf{0.27 $\pm$ 0.04} & $0.26 \pm 0.11$ & \textbf{1.0 $\pm$ 0.0} & $0.64 \pm 0.22$ \\
        \midrule
        DPO & $0.21 \pm 0.11$ & \textbf{0.01 $\pm$ 0.02} & \textbf{1.0 $\pm$ 0.0} & $0.54 \pm 0.06$ & $0.23 \pm 0.20$ & $0.01 \pm 0.01$ & \textbf{1.0 $\pm$ 0.0} & $0.78 \pm 0.28$ \\
        \midrule
        Probe & -- & $0.19 \pm 0.21$ & \textbf{1.0 $\pm$ 0.0} & \textbf{0.91 $\pm$ 0.13} & -- & $0.04 \pm 0.05$ & \textbf{1.0 $\pm$ 0.0} & $0.93 \pm 0.11$ \\
        \bottomrule
    \end{tabular}
    }
  }
  \caption{Evaluation when accepting multiple categories. We show it is quite possible to do so. In general SFT-based methods are best for in-domain performance, and CB or Probe are strong choices for OOD rejection.}
  \label{tab:multiple}
\end{table*}
\section{Experiments}\label{sec:exps}

In this section we explore a number of empirical questions: can we scope language models, how does scoping behave across scale, how much diversity is needed for scoping, or whether scoping is possible for multiple tasks simultaneously. We aim to be comprehensive, thus demonstrate results across 2-3 different categories per dataset. Where not detailed, our accept sets will be Sentiment Analysis (SA), Summarization (S) and Program Execution (PE).

All experiments contain evaluations of task performance (Accept Score) on the accept set (which should be high), rejection rate on the in-distribution accept (Accept) set (which should be low) as well as rejection rate on the in-distribution reject set (ID Reject) and out of distribution data (OOD Reject) (which should be high). We describe experiments in broad strokes, and defer precise details on hyperparameters to Appendix B. We begin by presenting results across a variety of language models (\texttt{Mistral-7B-Instruct-v0.2, granite-7b-instruct, Llama-3.1-8B-Instruct})~\citep{jiang2023mistral, sudalairaj2024lab, grattafiori2024llama}. We explore scale on the Llama-3.2-instruct family, and for other results choose Mistral-7b-instruct as we do not have the compute or space to run every experiment on every model.

\subsection{Scoping language models}\label{sec:scoping}

We start with the basic question: is it possible to scope language models? We explore each method across a variety of models using Sentiment Analysis (SA) as our accept task in Figure~\ref{fig:models}, where we see that although system prompting is insufficient, a broad variety of different methods are successful to different extents with different models.

\subsection{Rejection set diversity}\label{sec:diversity}

One of the most critical questions when attempting to restrict the generations of language models is what data might be necessary to do so. If models overfit to a particular data distribution, then it may be difficult to reject requests that were not specified in the training distribution. Thus, here we ask: how much data diversity is necessary in the rejection set to robustly scope models? If very little diversity is needed, and rejection extends to OOD requests, then adapting models to new deployments becomes quite inexpensive. In Figure~\ref{fig:diverse} we study Sentiment Analysis across different models with more and more diverse rejections sets consisting of an increasing number of tasks, and defer results on additional tasks with Mistral-7b-instruct to Appendix A.1.

We show results in Figure~\ref{fig:diverse}. In general we see that CB performs relatively better than SFT when data diversity is low, but SFT is much stronger with more rejection sets. Probing appears strong across the board, though it sometimes leads to overrejection on the Accept set, which is undesirable. In general it may be quite straightforward to collect diverse data, so unless the practitioner is quite constrained, it's worthwhile to always start with Probe and SFT.

\subsection{Accepting multiple tasks}\label{sec:multiple}

Here we ask: is it possible to still reject tasks when there are multiple tasks in the accept set? Such a setting is natural as most language models will have a few different specific uses, like a programming bot that can write code and also answer questions about documentation. We demonstrate results on \texttt{Mistral-7b-Instruct-v0.2} in Table~\ref{tab:multiple} with two choices of accepts sets: Classification and Generation (SA,S,TLD,SC,TC,DG) and Math and Programming (PE,GSM8k). We use the opposite set as the ID reject set in this case.

\begin{figure*}
  \centering
  \resizebox{0.9\textwidth}{!}{
  \begin{subfigure}[b]{0.23\linewidth}
    \centering
    \includegraphics[width=\linewidth]{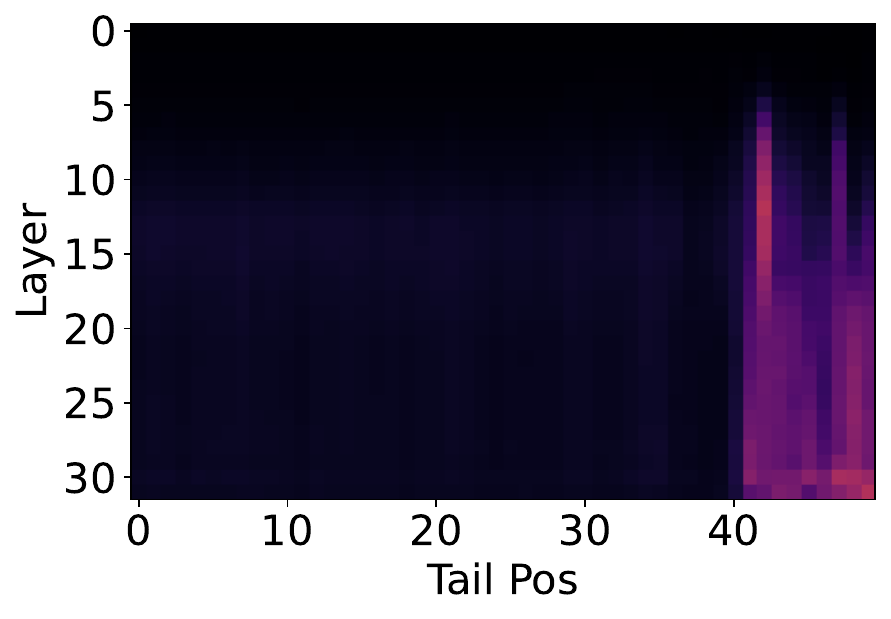}
    \\
    \includegraphics[width=\linewidth]{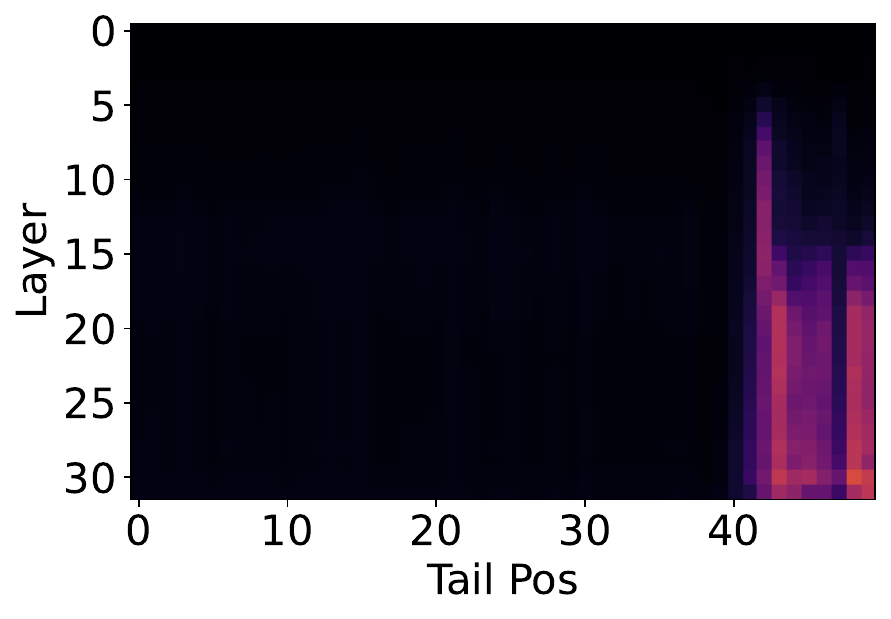}
    \\
    \includegraphics[width=\linewidth]{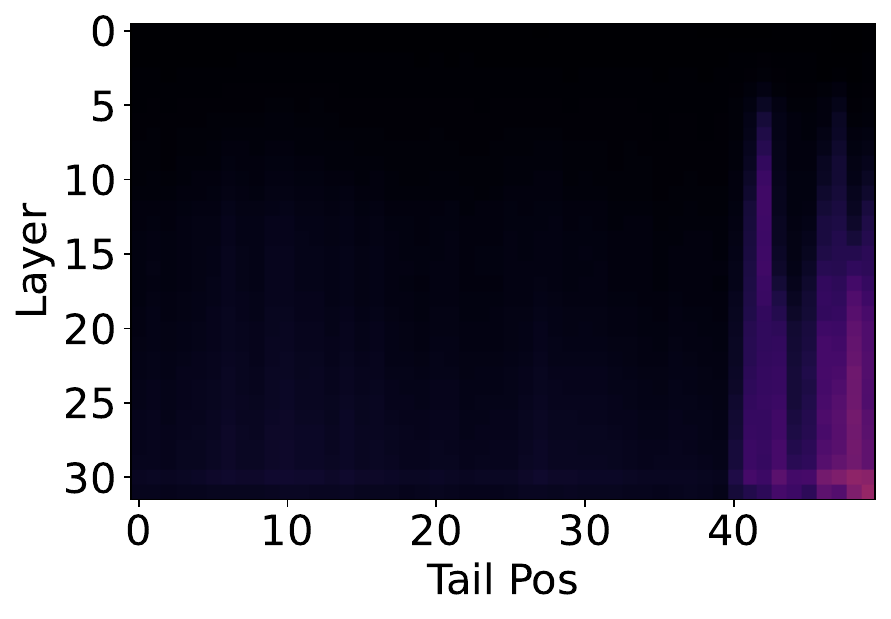}
    \caption{SFT}
  \end{subfigure}
  \begin{subfigure}[b]{0.23\linewidth}
    \centering
    \includegraphics[width=\linewidth]{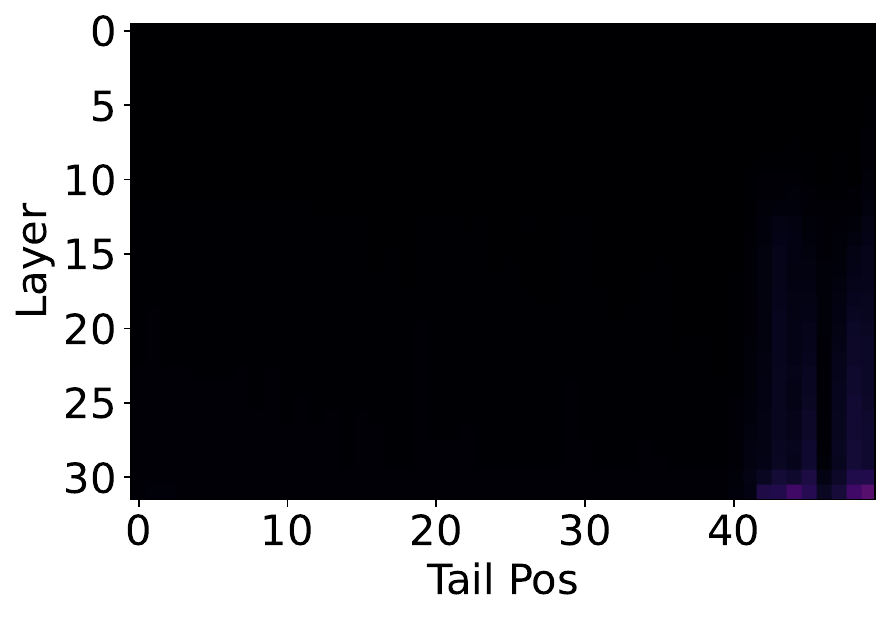}
    \\
    \includegraphics[width=\linewidth]{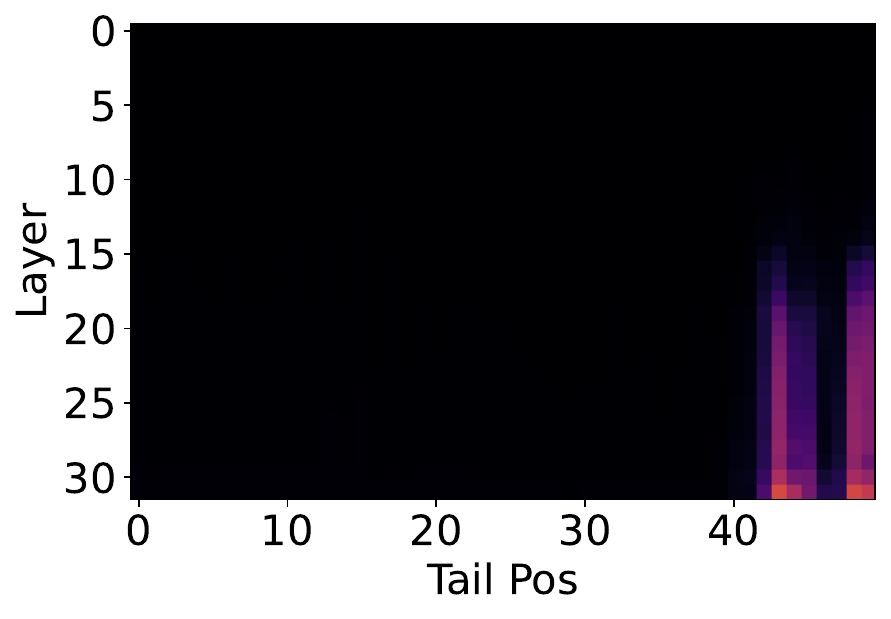}
    \\
    \includegraphics[width=\linewidth]{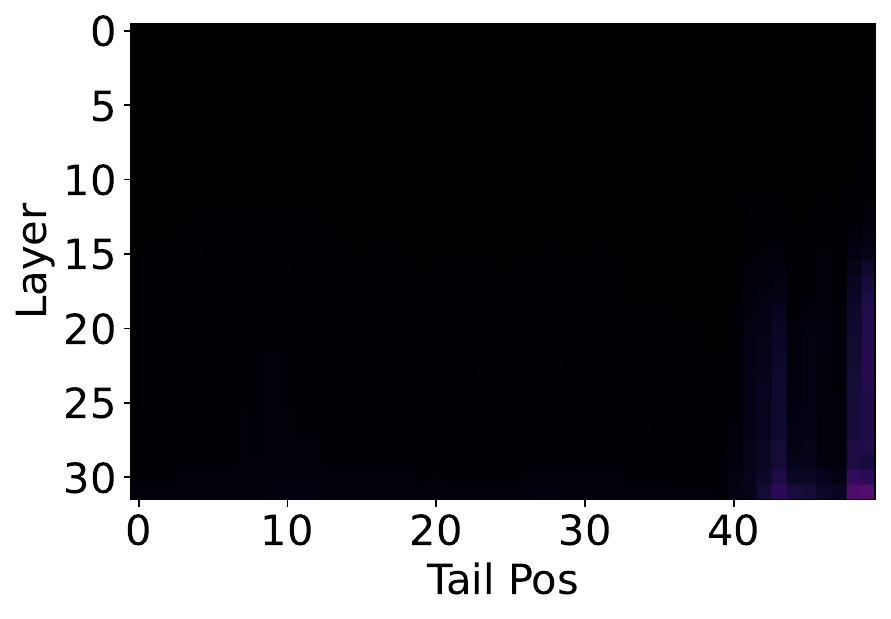}
    \caption{DPO}
  \end{subfigure}
  \begin{subfigure}[b]{0.23\linewidth}
    \centering
    \includegraphics[width=\linewidth]{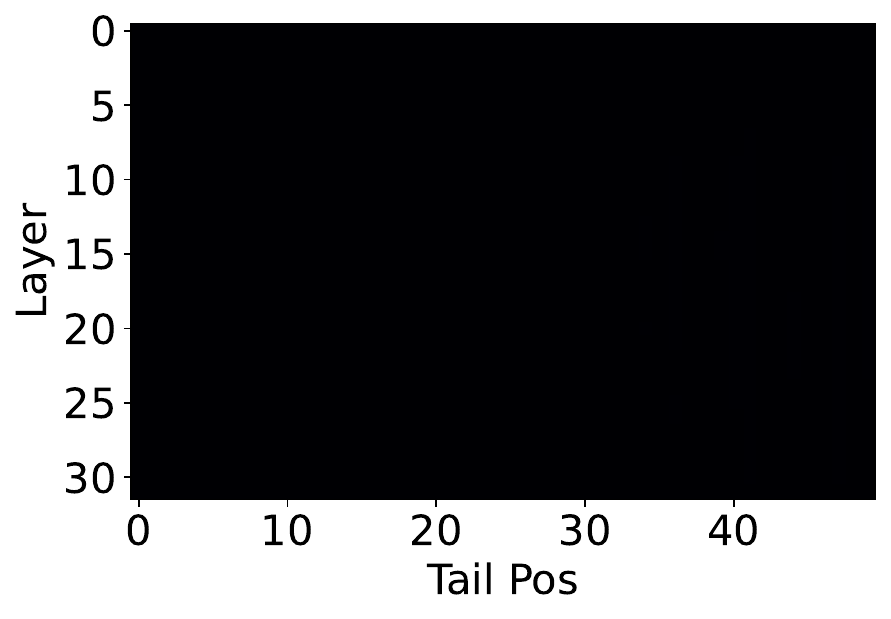}
    \\
    \includegraphics[width=\linewidth]{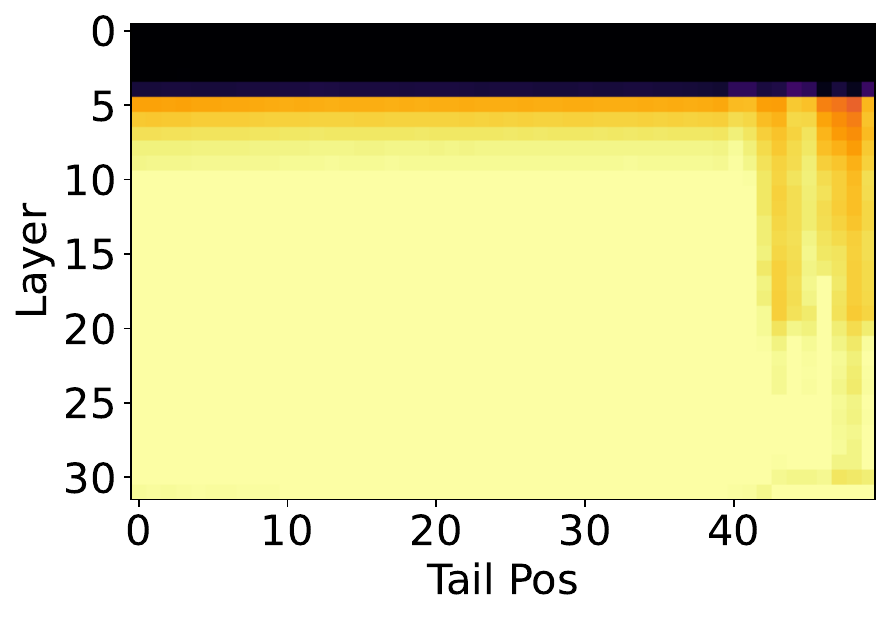}
    \\
    \includegraphics[width=\linewidth]{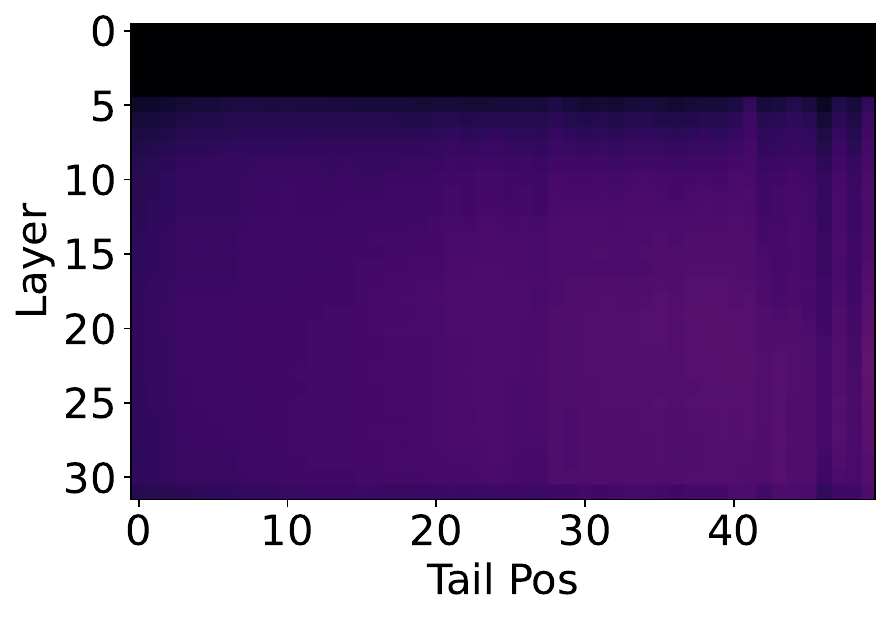}
    \caption{CB}
  \end{subfigure}
  \begin{subfigure}[b]{0.23\linewidth}
    \centering
    \includegraphics[width=\linewidth]{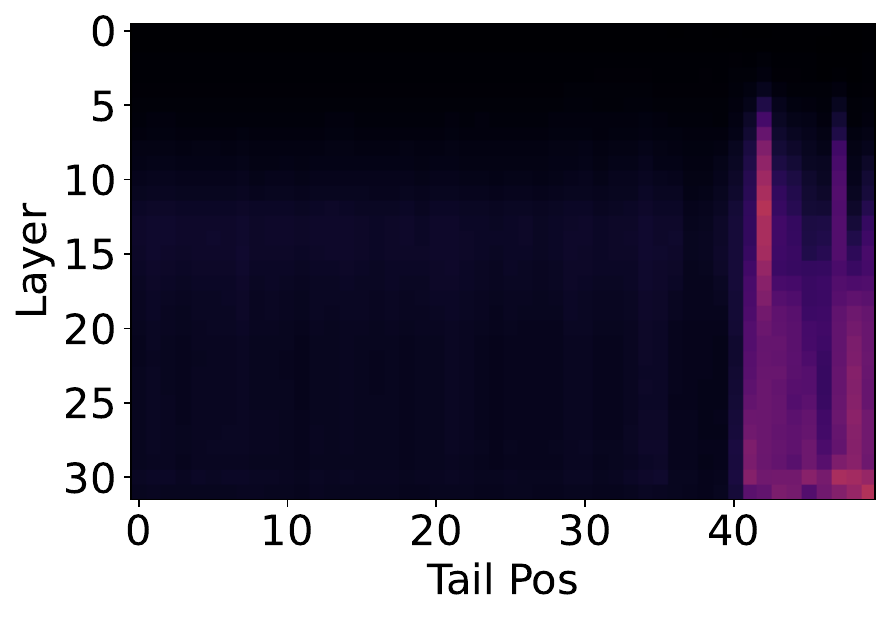}
    \\
    \includegraphics[width=\linewidth]{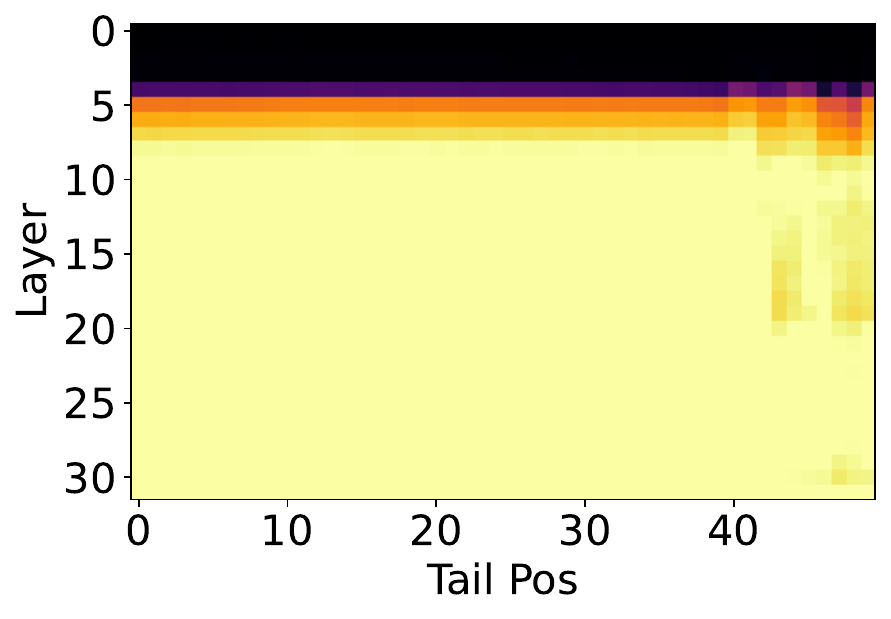}
    \\
    \includegraphics[width=\linewidth]{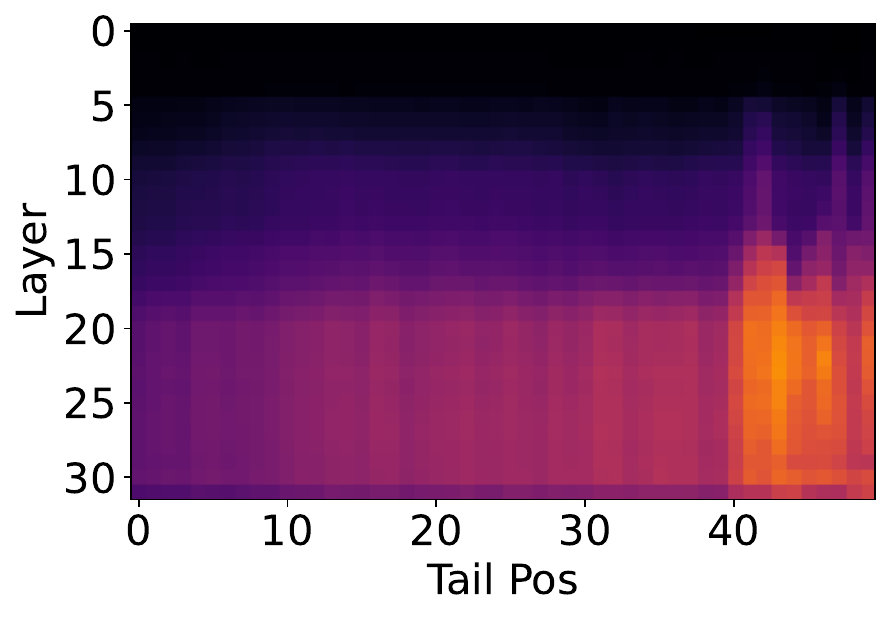}
    \caption{SFT $\rightarrow$ CB}
  \end{subfigure}
  \begin{subfigure}[b]{0.02\linewidth}
    \centering
    \includegraphics[height=25\linewidth]{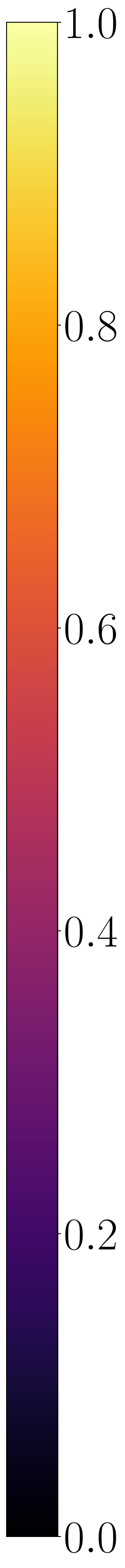}
    \caption*{}
  \end{subfigure}
  }
  \caption{Cosine distance between base model representations and LoRA tuned model. Rows represent model layer, columns represent tail position of prompt, averaged over all queries in the dataset. \textbf{Top row:} Accept data. \textbf{Middle row:} In distribution reject data. \textbf{Bottom row:} Out of distribution reject data. We see that CB-based methods make changes to representations across the context, while DPO and SFT only change the tail of the context. Layering SFT and CB yields both effects.}
  \label{fig:cosines}
\end{figure*}
\textbf{Classification and Generation:} We see strong scores for SFT-based methods here. On the accept set, Probe is worst, while Sys. is poor leading CB and SFT-CB to suffer. In distribution all methods work well except Sys. and SFT. Out of distribution, CB and Probe perform well, while SFT-CB and DPO are even.

\textbf{Math and Program Execution:} SFT-based methods perform best on the task. Surprisingly, SFT-CB has a very high rejection rate on the accept task. In-distribution every method but Sys. works well. Out-of-distribution there is a similar story to the previous case, where CB works quite well, and Probe is also strong, but the rest less so.

\textbf{Takeaways:} We see that it is possible to support multiple accept tasks. In particular, CB and Probe work best for out-of-distribution evaluation, but as its performance on the accept task is tied to the system prompt, any issues with the base model will carry over.

\subsection{Representation Analysis}\label{sec:cosine}

To get a sense for how different methods operate, we study how the representations change before and after training on \texttt{Mistral-7b-Instruct-v0.2}. In particular we look at cosine distance between the original representation of a token, and the same token after training. If there are patterns in these changes, it should give us some indication as to how different methods operate. In Figure~\ref{fig:cosines} we show that DPO and SFT only change the representations of the tail of the context. Hence it makes sense why CB is more robust under attack (explored in \citet{zou2024improving} and Appendix A.4): all representations have changed, so it is difficult to find a way to circumvent the changed behavior, while DPO and SFT have ``cracks" which can be exploited.

The effect is particularly clear on the in-distribution rejection set, but preceding sections demonstrate that most methods are fairly comparable in distribution. Out of distribution, the effect of CB is much less, though still there is a much more substantial difference from the original model than SFT or DPO which make only small changes to the tail of context in deeper layers. With SFT-CB, we can clearly see the layering of the tail edit as well as the orthogonalization across the entire context.

\subsection{Additional analysis}

Here we briefly discuss some additional results based on \texttt{Mistral-7b-Instruct-v0.2}, deferring full treatment to the Appendix.

\textbf{Adversarial Evaluation:} We examine the behavior of different methods under adversarial prompts in Appendix A.4, where we find that CB-based methods are more robust than others, echoing results of \citet{zou2024improving}.

\textbf{Model Scale:} In Appendix A.3,  we examine scoping behavior with different model scales, finding consistent results except for Probe which struggles with smaller models.

\textbf{Precise Scoping:} We find that one can scope precisely, (e.g. only News summarization instead of all summarization). In general many methods are appropriate for this, though consistent with prior results SFT suffers with a low diversity of training examples. For more details, see Appendix A.5.

\textbf{Tuning of CB:} In general, it was much more difficult to tune CB than SFT. Depending on the model, the optimal hyperparameters in terms of the target layers, and the choice of $\alpha$ made a large difference. This is important as CB often had very strong performance where appropriately tuned, but was less stable than other methods. We explore this more in Appendix A.6.

\textbf{Effect of Data Quantity:} We find that most methods work quite well with very little data (as little as 128 instances). DPO in particular benefits monotonically, while CB has issues as the raw number of training examples in the dataset scales, perhaps due to the difficulty of simultaneous orthogonalization of many different reject instances, see Appendix A.7 for more details.

\textbf{Effect of LoRA Rank:} Overall, it does appear that rank can have a substantial effect on the performance of methods. While DPO seems to scale monotonically with LoRA rank, CB-based methods have a sweet spot for performance, above which it seems optimization becomes difficult. See detailed analysis in Appendix A.8.

\section{Discussion}

Though current language models are generally applicable, there is still a need at deployment time to specify the kinds of queries they should and should not be able to answer. Otherwise, agents deployed in the wild may be easy to distract, and if used as a part of a pipeline may lead to cascading errors. Hence scoping is crucial. In this work, we conducted a comprehensive empirical study of scoping language models using three model families and multiple tasks.

Our findings reveal several key insights. First, system prompting alone is generally inadequate across a range of models and datasets. While performance varies depending on the specific method, model, and dataset, certain trends emerge. Supervised fine-tuning (SFT) tends to perform well when the training data is diverse, whereas Circuit Breakers (CB) is more effective in low-data regimes with less diversity, likely because the orthogonalization objective it employs is easier to optimize in such settings. Probing methods can also yield strong results, provided that the probe is sufficiently expressive to disentangle the model’s internal representations; however, this approach incurs additional inference overhead due to the use of an auxiliary model. Combining SFT and CB typically results in performance that reflects the strengths of both approaches, but this layered method is sensitive to the failure of either component, which can degrade overall performance.

We saw that it was possible to scope across language model scales, for multiple tasks at a time and for very fine-grained tasks. We demonstrated that different methods have different effects on the internal representations of the models: SFT and DPO only modify the tail of the language model context, unlike CB which modifies representations across the context. Such different behavior may explain why CB is stronger under adversarial attacks (Appendix A.4), the original setting it was proposed for~\citep{zou2024improving}. 

While these results indicate that CB can be a promising approach, it presents considerable challenges in practice. Specifically, its performance can fluctuate wildly based on very small step-count difference. Additionally, optimal target layers may need to be selected on a per-model basis, further complicating its application. Given these sensitivities, we recommend defaulting to simpler methods such as supervised fine-tuning (SFT) or probing, particularly in settings where data diversity is not a limiting factor.





\bibliography{aaai2026}

\appendix
\setcounter{secnumdepth}{2}            
\renewcommand{\thesection}{\Alph{section}}
\renewcommand{\thesubsection}{\thesection.\arabic{subsection}}

\section{Additional Results}

\subsection{Data diversity across different tasks}\label{app:diversity}

We continue discussion on data diversity. Here we examine performance with varying rejection set diversity using \texttt{Mistral-7b-Instruct-v0.2} on more tasks than just Sentiment Analysis.

\begin{figure*}[!h]
  \centering
  \begin{subfigure}[b]{0.5\linewidth}
    \centering
    \includegraphics[width=\linewidth]{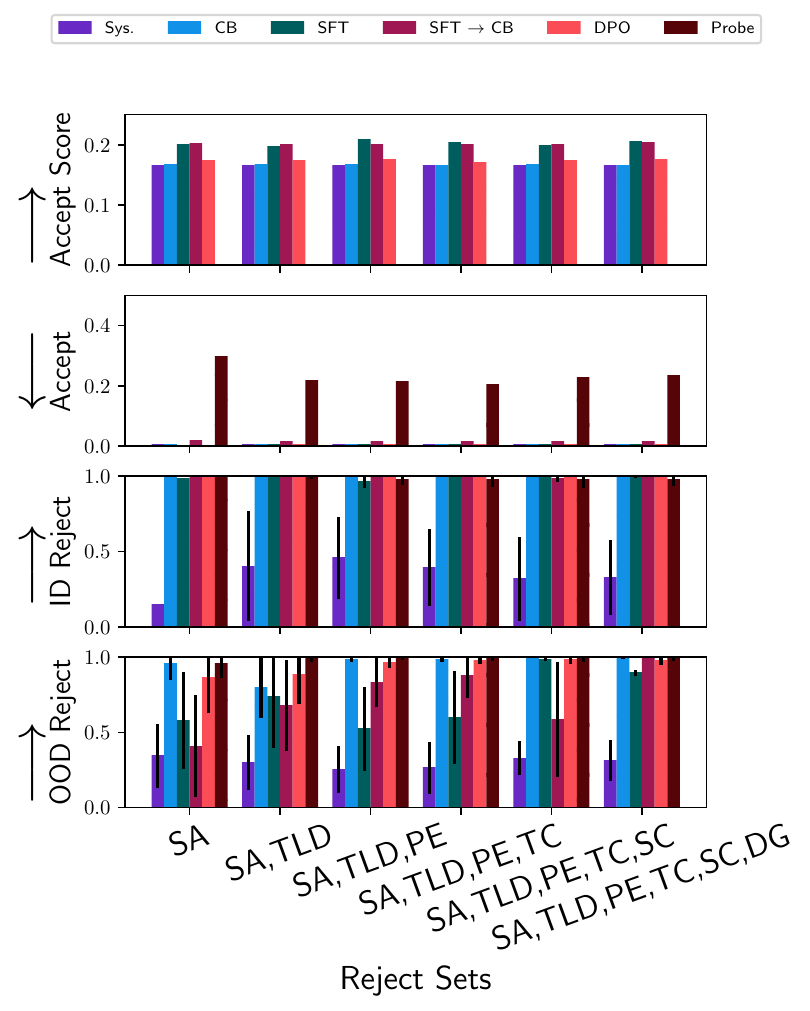}
    \caption{Summarization}
  \end{subfigure}\hfill\begin{subfigure}[b]{0.5\linewidth}
    \centering
    \includegraphics[width=\linewidth]{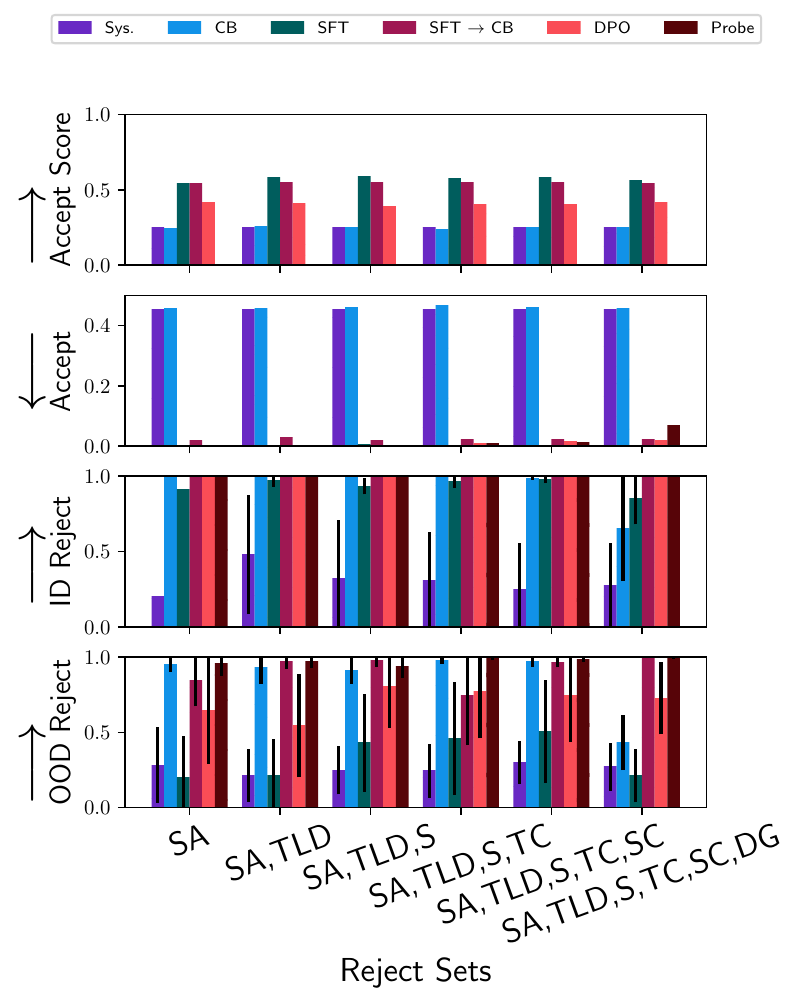}
    \caption{Program Execution}
  \end{subfigure}
  \caption{Results for increasing diversity of rejection set on tasks besides sentiment analysis (SA). Like in the SA case we see that CB performs worse with more diverse data.}
  \label{fig:task_diverse}
\end{figure*}

\textbf{Summarization:} In all cases, SFT based methods perform best on the task. Only Probe appears to reject the accept queries, with a rather high rate. In-distribution, Sys. is quite poor, but all other methods appear similar. Out-of-distribution, we see a slightly different story to classification, where CB is strong at low diversity, but so are DPO and Probe, while SFT-CB is not good until the data is quite diverse.

\textbf{Program Execution:} The same holds for task performance: SFT and SFT-CB are best. Sys. appears to reject this particular accept task at a high rate, and thus the CB rejection rate is also high. In-distribution there is not much trend as all methods except Sys. do well. Out of distribution we see that CB, SFT-CB and Probe are strong even when data diversity is poor, and similar to the Sentiment Analysis case, SFT-CB stays strong while CB suffers later.

\textbf{Takeaways:} At very low data diversity, CB and SFT-CB can still perform quite well. Probe also does well, though the rate of rejection on accept tasks can be high. As diversity increases, DPO becomes stronger and CB becomes weaker, though SFT-CB stays competitive.

\subsection{Additional Classification Baselines}

We previously compared the performance of scoping methods against a classifier (Probe) deployed in a two-stage (classify-then-generate) setup. For completeness, we compare the performance of the Probe Classifier against other baselines such as LLM judges based on LLama 3.1 8B instruct and Llama 3.2 70B Instruct deployed as classifiers, as well as a fine-tuned classifier based on a state-of-the-art general purpose text embedding based model - GTE 1.5. \footnote{https://huggingface.co/Alibaba-NLP/gte-base-en-v1.5} We train this model by adding a linear head for binary classification and tune all model parameters.

\begin{table}[ht]
\resizebox{.95\columnwidth}{!}{
\begin{tabular}{lccc}
\toprule
\textbf{Method} & \textbf{Accept} & \textbf{ID Reject} & \textbf{OOD Reject} \\
\midrule
 Probe Baseline         & 0.035 & $0.924 \pm 0.152$ & $0.952 \pm 0.107$ \\
Finetuned GTE     & 0.0   & 1.0              & $0.821 \pm 0.355$ \\
Llama 3.1 8B Instruct (Sys) & 	0.203 & $0.770 \pm 0.433$ & 	$0.801 \pm 0.146$ \\
Llama 3.1 8B Instruct (Judge)    & 0.082 & $0.549 \pm 0.162$ & $0.700 \pm 0.283$ \\
Llama 3.2 70B Instruct (Judge)  & 0.004 & $0.945 \pm 0.084$ & $0.819 \pm 0.358$ \\
\bottomrule
\end{tabular}
}
\caption{Classifier baselines}
\label{tab:Cls-results}
\end{table}

We see that prompted solutions at the same scale as the final model (8B, Sys. and Judge) are substantially worse both ID and OOD than the Probe baseline. We also see that at the 70B scale, the judge is still not better OOD. A large judge would not make sense at deployment time due to the high cost of inference. Thus we believe it is important to explore methods of LLM scoping.

For Finetuned GTE, the results are strong compared to many methods, but worse than the Probe baseline at the largest scale. Considering we will need to run the 8B model anyway, a GTE classifier solution is strictly more computationally inefficient and additionally more complex to manage.

\subsection{Scoping over different scales}\label{app:model_scale}

\begin{figure*}[!t]
  \centering
  \includegraphics[width=1.0\linewidth]{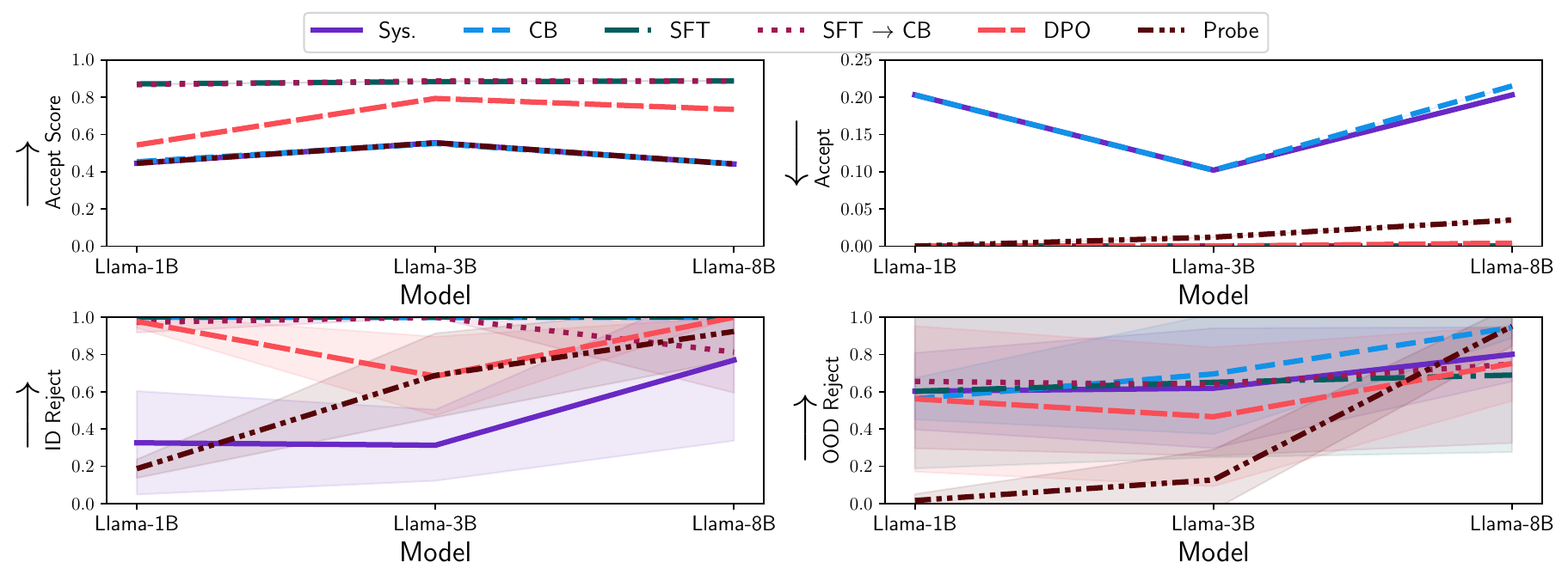}
  \caption{Scoping across different model scales. Larger models lead to improved results. In particular probing performs poorly at small scales, possibly due to the inability of a small probe to disentangle the representations well.}
  \label{fig:model_scale}
\end{figure*}

Here we ask: how does scoping vary with model scale? We use the Llama-3.1  family~\citep{grattafiori2024llama} as an example. We fix the accept task as Sentiment Analysis again and check how different methods behave with 1B, 3B and 8B parameter instruct-models. We see in Figure~\ref{fig:model_scale} broadly that larger models lead to improved results. The only other major surprise is that probing performs poorly with smaller language models. This may be due to the fact that it is difficult for a small probe to disentangle the representations of smaller models, as they are lower dimension and the information is more compressed.

\subsection{Robustness to Adversarial Prompts}\label{sec:adv_eval}

The original CB methodology~\citep{zou2024improving}, and many related works discussed, focus on robustness to adversarial prompts. If models are to be deployed, we might expect that users could attempt adversarial attacks against the deployment. Here we ask: if models are scoped, how robust are they to adversarial prompts? We take \texttt{Mistral-7b-Instruct-v0.2} as a case study here.

Our threat model is of black-box access. In particular, we assume that the users are allowed to edit only the instruction text, and do not even have access to modifying the system prompt, which would be true for text-based API access. We implement and test a number of different black-box adversarial attacks:

\textbf{Adversarial system prompt (Adv.):} We insert an adversarial system prompt at the beginning of the instruction, after the original system prompt. This adversarial system prompt is of the same format as the original, but instead of being for the category of the training accept task, it corresponds to the category of the rejection task.

\textbf{Base-64 translation (b64):} Following \citet{wei2024jailbroken}, we translate instructions into base-64, then prompt the language model. After receiving the response if it is valid base-64 (which is very often), we translate it back to text.

\textbf{Few-shot prompting (Few-shot):} We provide a few-shot example from the evaluation set, where we draw a training query and completion and then prompt the next round with a new query. This is similar to the Many-Shot attack explored by ~\citet{anil2024many}.

\textbf{Multiturn prompting (2-turn):} We prompt with a full conversation turn of an accept task and accept completion, then a second turn with a rejection request from the reject set. This format intends to prime the model to first get into an "accept" mode, before answering the new query.

\textbf{Multiturn prompting with adversarial system prompt (2-turn+Sys.):} This is similar to the attack above, but we add an adversarial system prompt to the beginning of the 2nd turn.

\textbf{Prefill:} In this attack, append a generic prefilling output (\textit{"Sure, here's a response to your request:"}), to the end of the user instruction. This follows common practice and has been a strong attack~\citep{wei2024jailbroken, zou2024improving}.

\textbf{Tree of Attacks with Pruning (TAP):} TAP~\citep{mehrotra2023tree} is an adversarial prompting method whereby an attacking language model iteratively attempts to jailbreak a target language model (here our model to evaluate). TAP uses an actor-critic loop to rewrite prompts based on whether the language model was jailbroken in the previous step, and explores a tree to find the best performing prompt. For more details and hyperparameters, see Appendix of Experimental Details. It is a strong black-box optimization-based jailbreaking method, and was among the strongest attacks for CB in the original setting~\citep{zou2024improving}. As TAP is quite expensive to run, we only test 10 prompts per dataset.

\begin{figure*}
  \centering
  \begin{subfigure}[b]{0.3333\textwidth}
    \centering
    \includegraphics[width=\linewidth]{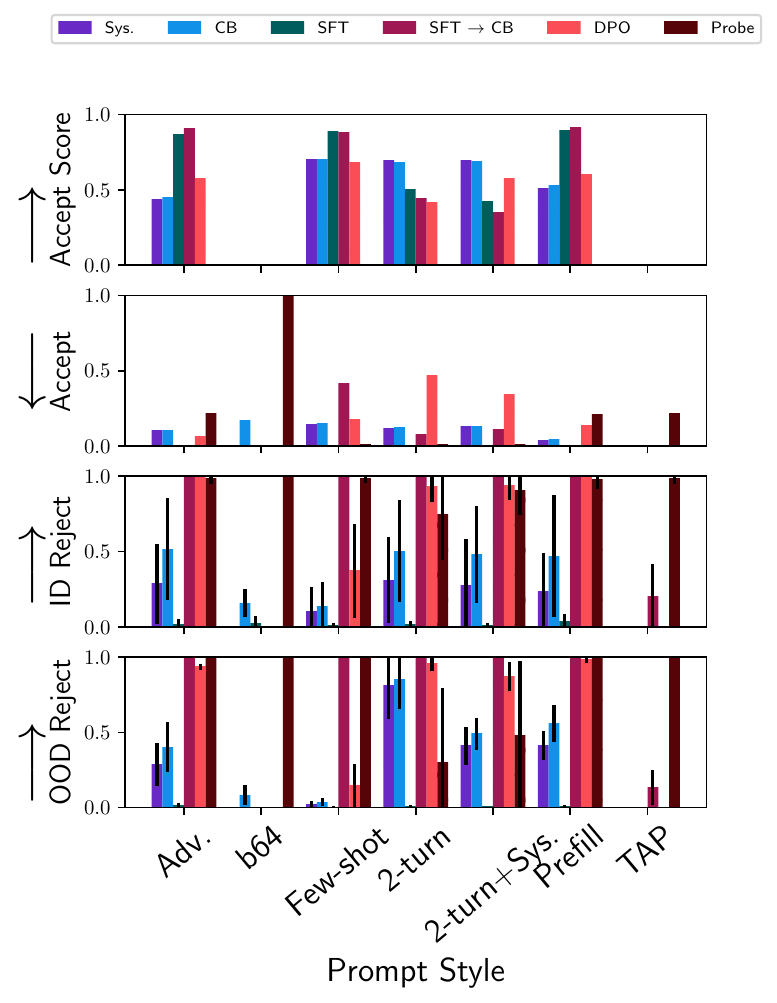}
    \caption{Sentiment Analysis}
  \end{subfigure}\hfill\begin{subfigure}[b]{0.3333\linewidth}
    \centering
    \includegraphics[width=\linewidth]{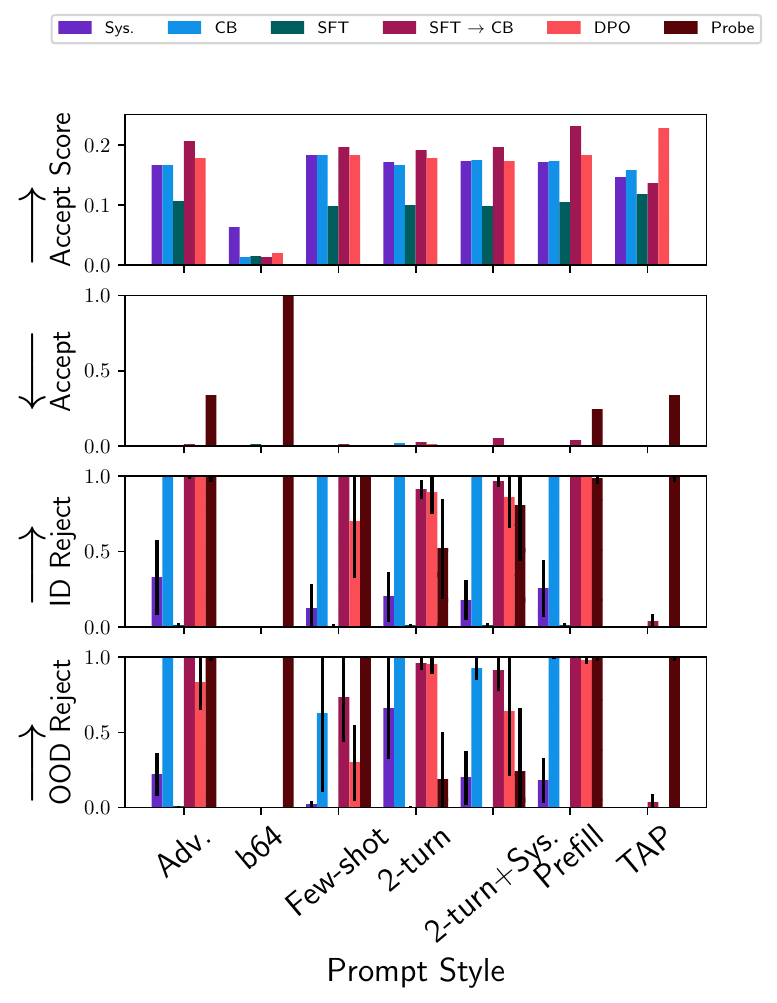}
    \caption{Summarization}
  \end{subfigure}\hfill\begin{subfigure}[b]{0.3333\linewidth}
    \centering
    \includegraphics[width=\linewidth]{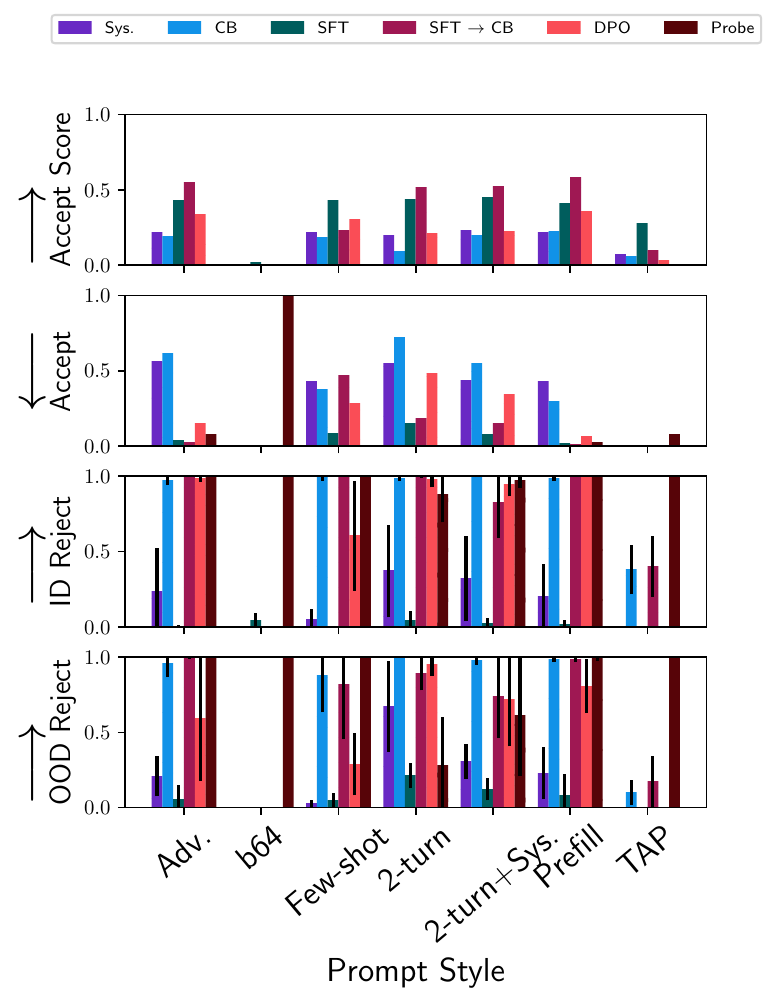}
    \caption{Program Execution}
  \end{subfigure}
  \caption{Robustness evaluation for Mistral.}
  \label{fig:adv_eval}
\end{figure*}

We show results for all evaluations in Figure~\ref{fig:adv_eval}.

\textbf{Sentiment Analysis:} As far as performance, we see that CB and SFT-CB are very similar to Sys. and SFT respectively. SFT-based methods perform best except when distractor turns are added, which may be due to a mismatch between training and evaluation. The rejection rate on the accept task is very low, though the Probe and DPO seem to have a tendency toward over-rejection. In-distribution, both the Probe and SFT-CB seem to perform very well, with DPO in 3rd. Out of distribution there is a similar trend, though the probe suffers from the 2-turn attack. When subject to the strong iterative prompting attack the Probe is best.

\textbf{Summarization:} Here SFT-CB performs best among all methods, except under TAP prompting where DPO is better. While the Probe has a tendency toward over-rejection on the accept set, all otehr methods perform well. In-distribution, CB and SFT-CB and Probe are strong, while DPO suffers. Out of distribution we see a similar trend. When SFT-CB does poorly, CB itself is still strong.

\textbf{Program Execution:} SFT and SFT-CB are the strongest on task performance in all cases. Rejection rates on the accept set are high for the untuned language model (Sys.), hence also for CB which preserves the function. DPO also shows a tendency to reject in multiple cases. In-distribution CB, SFT-CB and Probe are again strongest, with PO trailing. Out-of-distribution the case where DPO beats SFT-CB (2-turn), it is quite close, and CB is near perfect.

\textbf{Takeaways:} Notably, both Sys. and SFT are quite poor. DPO as well has many issues. Probe is quite strong, but also has a tendency to reject accept tasks, which is undesirable. Thus it appears CB, and SFT-CB, strike a nice balance between in and out of distribution rejection, as well as letting desired prompts pass through. In all of these evaluations it is clear that none of these methods are even close to perfect, so there is still much work to be done. Such results are quite distinct from the safety picture presented by \citet{zou2024improving}, perhaps as the domains are not quite as simple as safe vs. unsafe prompts. Still, the spirit of the results in \citet{zou2024improving} appear to be true: CB seems more robust to adversarial attacks than baselines, with the exception of the Probe which tends to reject even on accept tasks. One additional point on the b64 attack, which appears to bypass all models: the completions tend to either be generic base-64 encoded response (e.g. \textit{"Hello world!"}), or invalid base-64.

\subsection{Precise Scoping}\label{sec:precise}

\begin{figure*}
  \centering
  \begin{subfigure}[b]{0.49\linewidth}
    \centering
    \includegraphics[width=\linewidth]{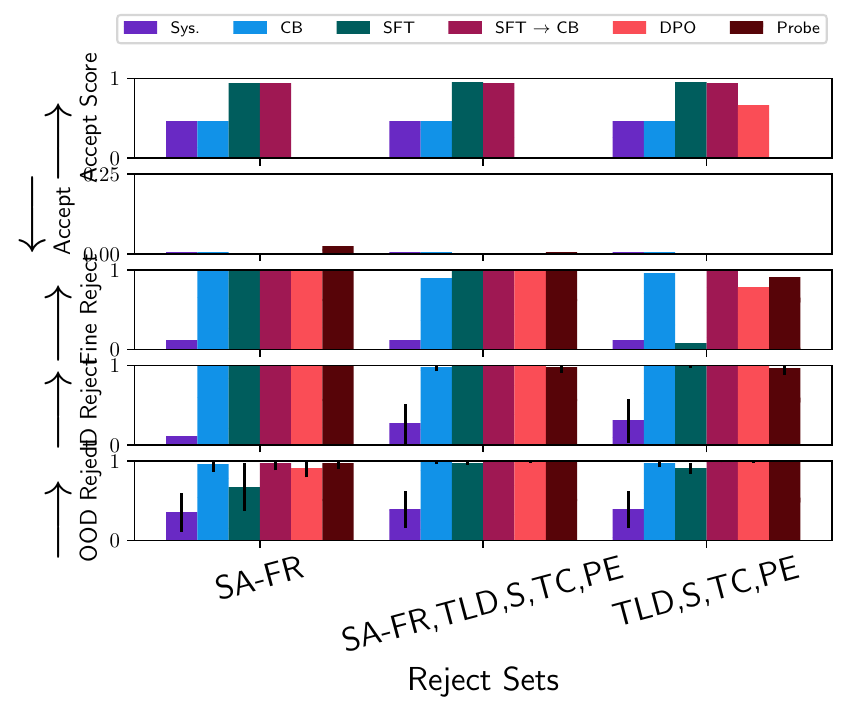}
    \caption{Sentiment Analysis}
  \end{subfigure}
  \hfill
  \begin{subfigure}[b]{0.49\linewidth}
    \centering
    \includegraphics[width=\linewidth]{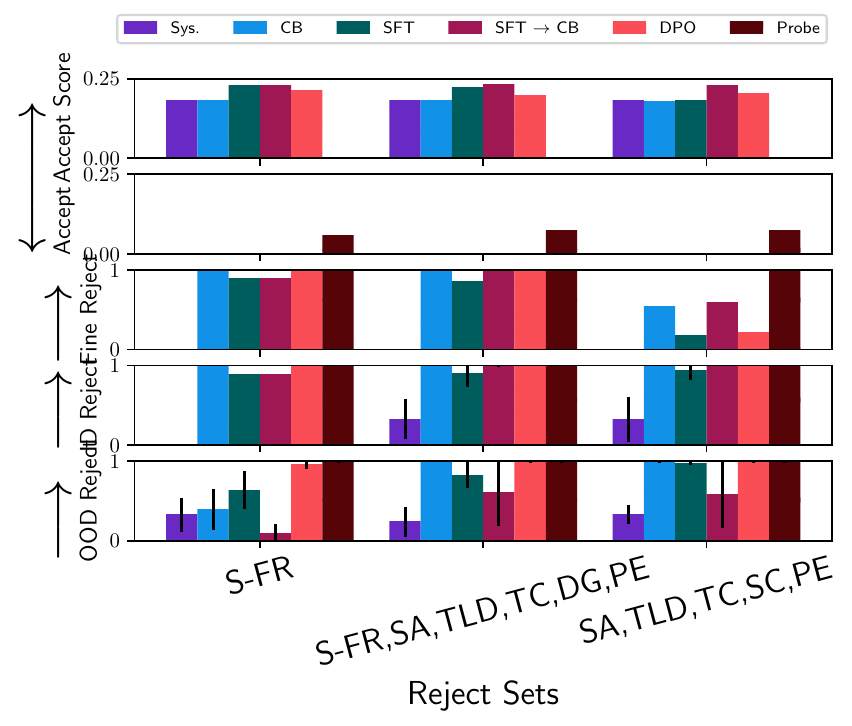}
    \caption{Summarization}
  \end{subfigure}
  \caption{Results for scoping on precise tasks.}
  \label{fig:precise}
\end{figure*}

Here we ask the question: how precisely can you scope? As an example, is it possible to scope not only to summarization in general, but \textit{only} to news summarization, rejecting all other requests including summarization ones. Here we create a fine-grained accept (FA) and fine-grained reject (FR) set from a categories of tasks like SA by holding one single task within that category as SA-FA, and taking all the rest as SA-FR. We do similarly for summarization. We show results in Figure~\ref{fig:precise} using \texttt{Mistral-7b-Instruct-v0.2} as a case study.

\textbf{Sentiment Analysis:} For task performance, unsurprisingly SFT-based methods are best. Strangely DPO seems to suffer when SA-FR is included in the rejection set. All methods have no rejections on the accept task. For the fine-grained rejection set, all methods do well (except Sys.) when it is included in the rejection set, but CB-based methods do best when it is not (see last column). On in-distribution rejection, all methods do well. For out of distribution, we see that CB, SFT-CB and Probe are best on the low-diversity case (only SA-FR), while as the distribution expands other methods catch up, echoing previous results.

\textbf{Summarization:} For task performance we see a consistent story with other plots. On the accept set, only Probe has any rejections. Similar to the previous case, when S-FR is not included in the rejection set, CB, SFT-CB and Probe do well, but other methods do not, however when it is included DPO is also very strong. In-distribution there is not much difference between methods. Out of distribution, when the data distribution is very narrow surprisingly both CB and SFT-CB are very poor. DPO, however, does quite well. As the data distribution expands, CB does better, but SFT-CB is still poor.

\textbf{Takeaways:} First it does appear to be the case that fine-grained scoping is possible. It is difficult to decisively say one method is best given the differences between the two tasks, and all methods appear to perform well when the fine-grained rejection set is provided for training. However, we do see that SFT-CB, CB and Probe can do well even when the fine-grained rejection set is not provided for training.

\subsection{Ablations on Circuit Breaker hyperparameters}\label{sec:cb_hyps}

\begin{figure*}[!h]
  \centering
  \includegraphics[width=\linewidth]{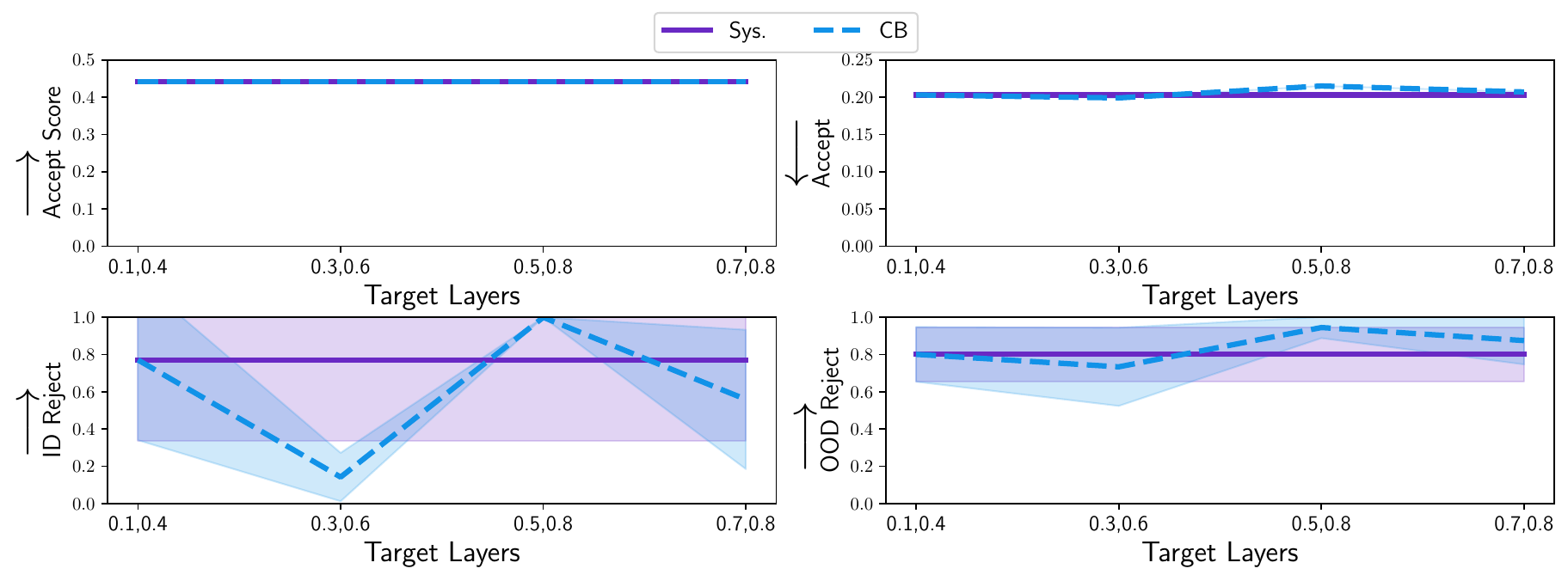}
  \caption{Results tuning CB with different target layers on \texttt{Llama-3.1-8B-Instruct}.}
  \label{fig:cb_target_layer}
\end{figure*}

\begin{figure*}[!h]
  \centering
  \includegraphics[width=\linewidth]{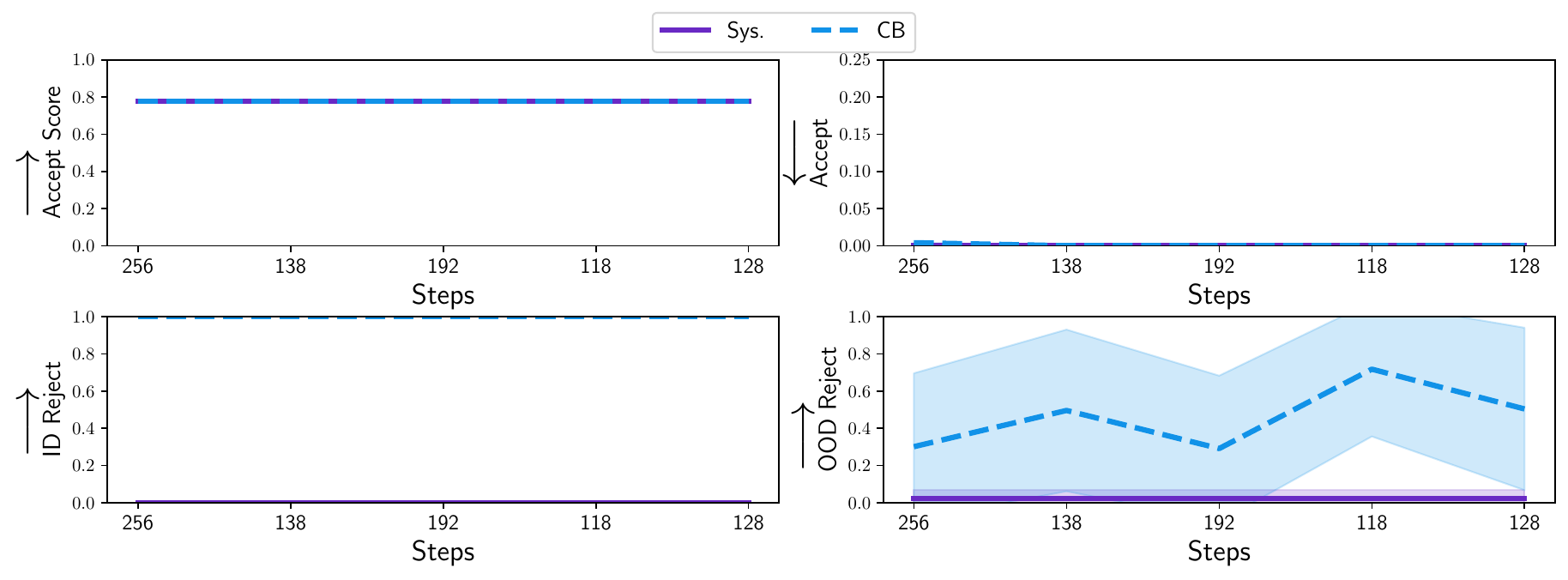}
  \caption{Results tuning CB with different number of steps with \texttt{Granite-7b-Instruct}.}
  \label{fig:cb_num_steps}
\end{figure*}

CB is much more difficult to tune than other methods, across models. Depending on the model, the optimal choice of the target layers (Figure~\ref{fig:cb_target_layer}), the layers to orthogonalize, and the choice of step budget make a large difference. One might expect large differences in target layer choices because different models could represent different patterns relevant to the task at different layers, so it might be easier or harder to orthogonalize based on that choice. The fact that performance fluctuates so much with step count (Figure~\ref{fig:cb_num_steps}), however, is quite strange when compared to other methods like SFT or DPO, and led to much more difficult tuning cycles.

\subsection{Effect of Data Quantity}\label{sec:quantity}

\begin{figure*}[!h]
  \centering
  \begin{subfigure}[b]{0.32\linewidth}
    \centering
    \includegraphics[width=\linewidth]{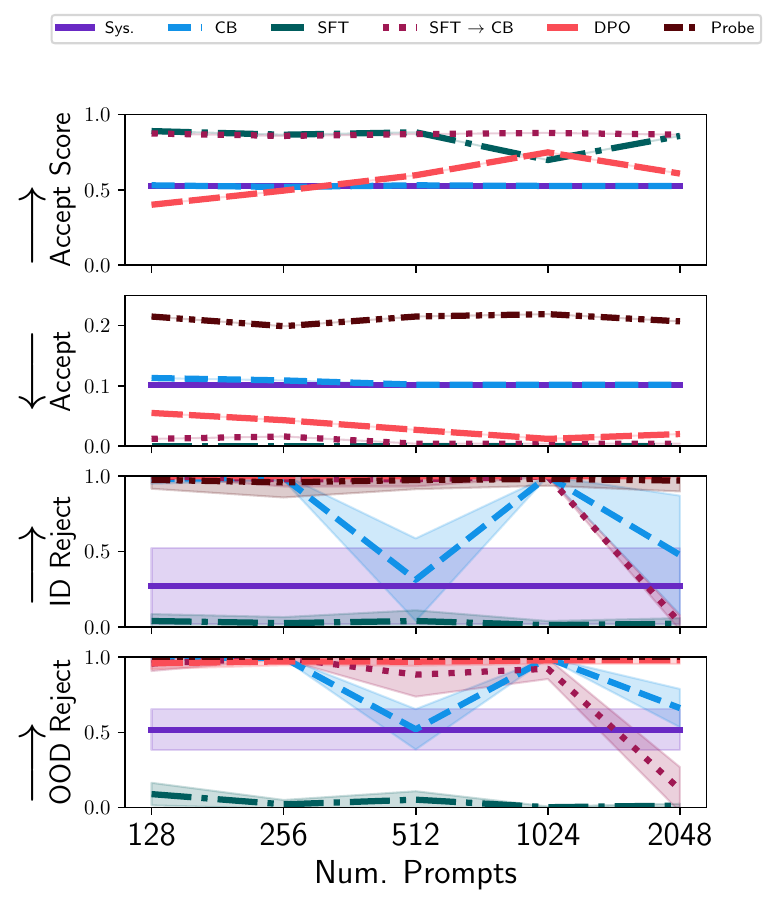}
    \caption{Sentiment Analysis}
  \end{subfigure}
  \hfill
  \begin{subfigure}[b]{0.32\linewidth}
    \centering
    \includegraphics[width=\linewidth]{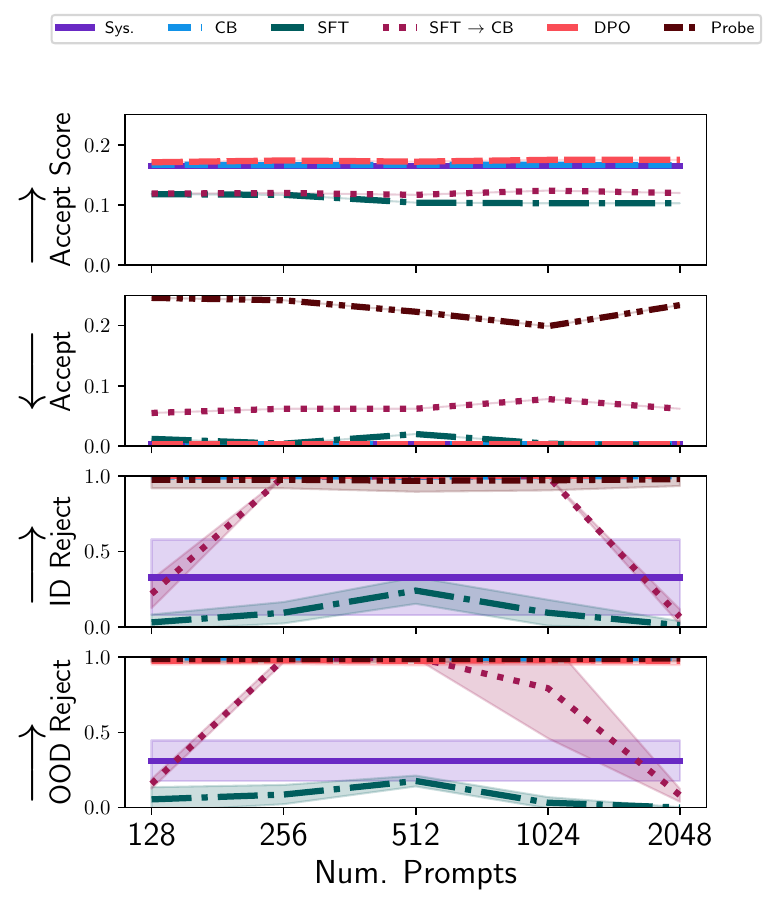}
    \caption{Summarization}
  \end{subfigure}
  \hfill
  \begin{subfigure}[b]{0.32\linewidth}
    \centering
    \includegraphics[width=\linewidth]{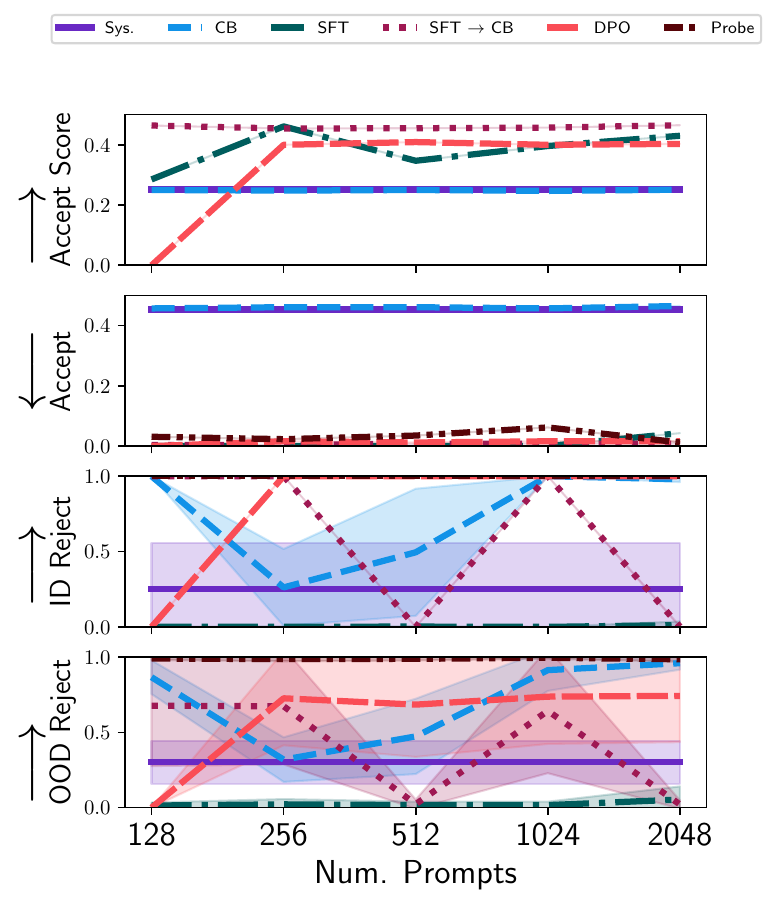}
    \caption{Program Execution}
  \end{subfigure}
  \caption{Evaluations with increasing number of instances in the accept and reject sets.}
  \label{fig:quantity}
\end{figure*}

Here we wonder: how important is the quantity of instructions in accept and reject sets? It would be ideal if only very little data were needed to learn the desired behavior, as it would make spinning up new deployments very speedy. We demonstrate all evaluations in Figure~\ref{fig:quantity} on \texttt{Mistral-7b-Instruct-v0.2}.

\textbf{Sentiment Analysis:} Perhaps unsurprisingly, SFT-based methods are best across the board. Interestingly, very little data is needed for this task and scores are roughly flat. On the accept set, rejection rates are also flat with the number of prompts, and the Probe always rejects a large number. In-distribution, the major trend to note is that both DPO and Probe are quite stable and strong across number of prompts, but CB appears quite unstable and seesaws. This may be due to difficulty optimizing for orthogonality. A similar trend is visible in the OOD case.

\textbf{Summarization:} DPO appears best here in terms of task performance. Trends are flat and Probe is worst on the accept task rejection rate. For ID reject SFT-based methods seem to have a hump structure, doing best in the middle of the range, and similarly for OOD.

\textbf{Program Execution:} Here SFT-CB and DPO perform best, though DPO requires more data to perform well. Both CB and Sys. have high rejection rates on accept due to base language model behavior. Both the in-distribution and out-of-distribution plots are quite noisy, so it is difficult to draw any strong conclusions besides the fact that the Probe does well.

\textbf{Takeaways:} It appears that the Probe is the most stable of methods for all amounts of data. Among the different tasks there is a significant amount of variability between methods, so it is difficult to make general comments. It is true, however, that some methods in each case work with very little data.

\subsection{Effect of LoRA Rank} \label{sec:rank}

\begin{figure*}[!h]
  \centering
  \begin{subfigure}[b]{0.32\linewidth}
    \centering
    \includegraphics[width=\linewidth]{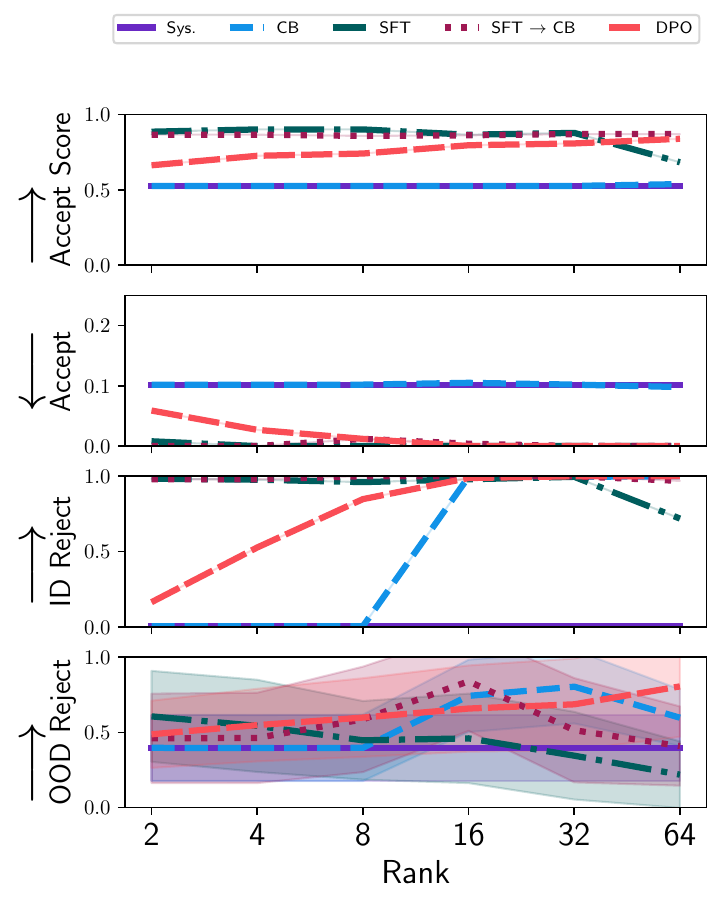}
    \caption{Sentiment Analysis}
  \end{subfigure}
  \hfill
  \begin{subfigure}[b]{0.32\linewidth}
    \centering
    \includegraphics[width=\linewidth]{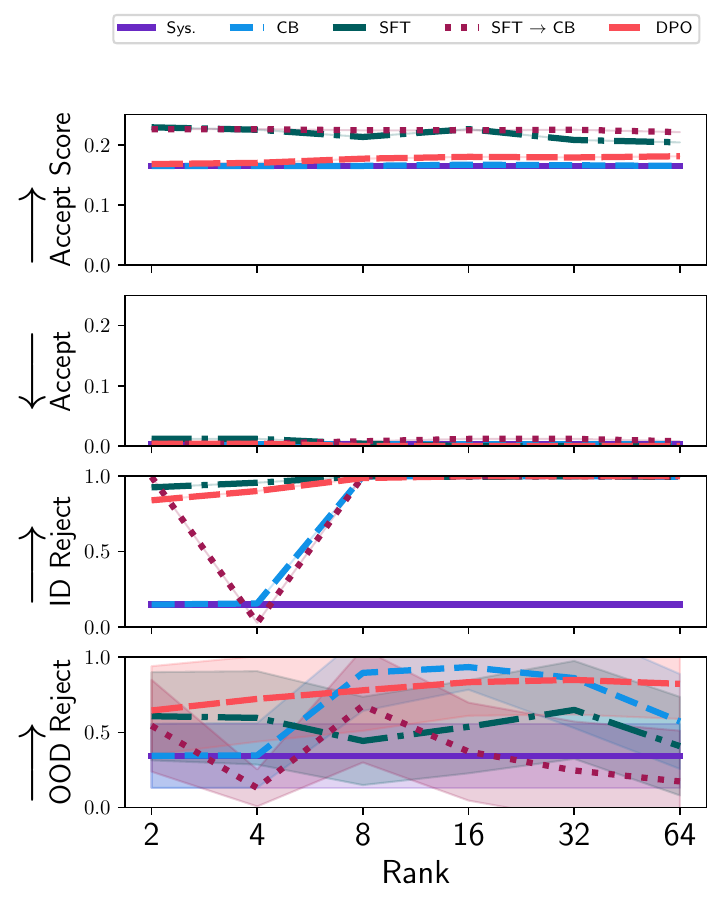}
    \caption{Summarization}
  \end{subfigure}
  \hfill
  \begin{subfigure}[b]{0.32\linewidth}
    \centering
    \includegraphics[width=\linewidth]{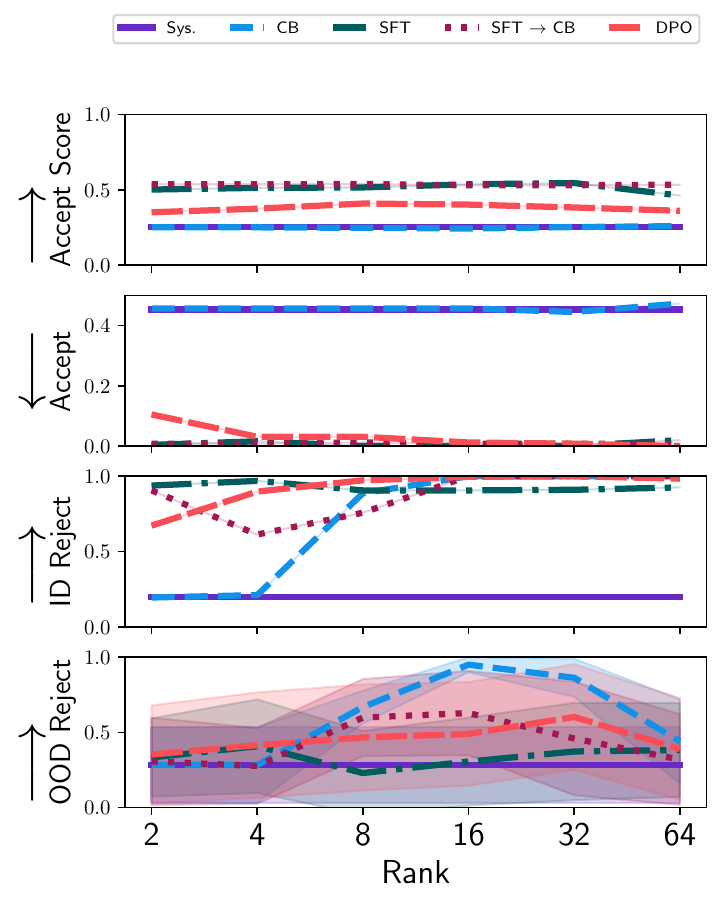}
    \caption{Program Execution}
  \end{subfigure}
  \caption{Results for increasing LoRA rank.}
  \label{fig:rank}
\end{figure*}

All methods except Probe rely on LoRA. Here we ask: is there a benefit to additional LoRA capacity, as expressed in the rank? It might be logical to expect that different tasks would have a different optimal rank, and we study that below. Our findings are shown in Figure~\ref{fig:rank} on \texttt{Mistral-7b-Instruct-v0.2}.

\textbf{Sentiment Analysis:} The performance and rejection rates of DPO both appear to increase monotonically with rank, but for other methods the trend is unclear. SFT-CB in particular is largely flat except for the OOD performance, which is best in the middle. This might be because it is difficult to optimize orthogonality in so many dimensions, but relatively straightforward in fewer.

\textbf{Summarization:} Here again there is a very slight monotonic trend with rank for DPO, but for other methods we do not see such trends. CB seems better at the higher end, and performs best of all methods OOD, but as rank reaches its maximum CB does worse.

\textbf{Program Execution:} Once again we see a similar story, though a large gap between the best CB setting OOD and the reset of the methods.

\textbf{Takeaways:} Overall, it does appear that rank is important and can have a substantial effect on the performance of methods. While DPO seems to scale monotonically with LoRA rank, CB-based methods have a sweet spot for performance, above which it seems optimization becomes difficult.

\section{Experimental Details}\label{sec:exp_details}

In the following sections we provide details on experimental hyperparameters for clarity.

\subsection{Training}

For CB, we follow \citet{zou2024improving} and add LoRA to all matrix parameters (both Attention and MLP), but only on the first 20 layers. We use layers 10 and 20 as targets for the representations, as described by \citet{zou2024improving}. For SFT and DPO we add LoRA to all matrix parameters for all 32 layers. For all experiments we use Adam~\citep{kingma2014adam} without weight decay.

Along with details previously described in the main text, we tune all methods for a single set of accept and reject sets (SA vs. S, TC, SC, DG) with a learning rate grid of $\{1\cdot 10^{-6}, 5\cdot 10^{-6}, 1\cdot 10^{-5}, 5\cdot 10^{-5}, 1\cdot 10^{-4}, 5\cdot 10^{-4}, 1\cdot 10^{-3}\}$, and step grid of $\{64, 128, 256, 512, 1024\}$. For DPO, we also tune the KL regularization coefficient $\beta$ in a grid $\{0.05, 0.1, 0.25, 0.5, 0.75\}$. This results in the following default configurations:

\begin{itemize}[itemsep=1pt,topsep=0pt,parsep=0pt,partopsep=0pt]
    \item \textbf{CB:} LR 0.0001, 128 steps.
    \item \textbf{SFT:} LR 0.0001, 256 steps.
    \item \textbf{SFT-CB:} the two above in sequence.
    \item \textbf{DPO:} LR 0.00001, 512 steps.
    \item \textbf{Probe:} LR 0.00001, 256 steps.
\end{itemize}

\subsection{Evaluation}

For every category, when we evaluate we use 256 heldout prompts. The only exception is for TAP~\citep{mehrotra2023tree}, where due to high costs we only evaluate 10 prompts per category.

Ideally one might choose to use a language model judge for detecting rejection~\citep{zheng2023judging}. However, given the large number of experiments and evaluations in this work, we found it prohibitively expensive to run all the evaluations through a state-of-the-art API judge. We experimented with using locally hosted language models as judges, with the largest being \texttt{Llama-3-70B-Instruct}~\citep{dubey2024llama}, but found such detection to have much poorer performance both in precision and recall than the methods described below.

As different methods behave differently, we employ different ways to detect rejection. For all methods besides probing, as the system prompt and tuning will instruct models to respond ``I cannot answer that.'', we catch rejection by string matching for a few different tokens that are synonyms for ``cannot'' at the beginning of the generation. The reason we only match strings early in the generation is that it is possible to switch from reject to accept behavior, so we would like to catch that rejection early on. We do not match strings later on as in practice we never observed language models switching from accept to reject behavior midway through generation and keywords can be used as a part of a legitimate response later in generation. On a sample of 30 completions from accept, reject, and OOD reject sets, we tuned the threshold that such a detector had perfect agreement with manual judgment. This was inspired by common string-based detectors like the one proposed by \citet{zou2023universal} and used by \citet{zeng2024johnny, zou2024improving}.

The full list of strings that we match for the string-based detector is: ["cannot", "unable", "sorry", "can't", "will not", "won't", "do not", "don't", "not able"]. We arrived at this list after manual verification of a subset of 90 prompts, and we find the performance to match human judgments. As we are tuning, we are unable to simply reuse the list from \citet{zou2023universal} as model completions may change. If we detect any of these words within the first 30 characters (the beginning of the completion), we consider the model to reject that prompt.

For the CB rejection detection, generations break often by repeating the same token over and over again. As exact-matching does not detect such patterns, in addition to string matching described above which will activate when the system prompt is followed, we catch rejection by the existence of a repeated pattern of 4 or more strings within the response. Again on a sample of 30 completions form accept, reject and OOD reject sets, we tuned the threshold for this detector such that it achieved 1 false negative and 0 false positives out of 90 completions. The single false negative was due to to a broken generation of punctuation characters that lacked repetitions. See Appendix of Sample outputs.

For Probing, we simply use the binary classification decision from the MLP as the rejection decision.

\subsection{Scoping language models}

In Section of Scoping language models, we use \texttt{Mistral-7B-Instruct-v0.2}, \texttt{Granite-7b-Instruct} and \texttt{Llama-3.1-8B-Instruct} for results. Our accept set is SA while the reject set is S, TLD, TC, SC.

\subsection{Rejection Set Diversity}

In order to be consistent between experiments, we keep the same number of instances in accept and reject sets (2048). Thus as the rejection set grows more diverse, there are fewer instances per category. It appears this does not have a significant effect on methods like DPO, so we believe this quantitative decrease should not have any major downsides.

For the results in Figure~\ref{fig:diverse}, we use \texttt{Mistral-7B-Instruct-v0.2}, \texttt{Granite-7b-Instruct} and \texttt{Llama-3.2-8B-Instruct}. We use SA as our accept set and S, TLD, TC, SC as our reject set.

For Figure~\ref{fig:diverse} we fix the model to \texttt{Mistral-7B-Instruct-v0.2} and use S and PE as the two accept sets.

\begin{figure*}[!t]
  \centering
  \includegraphics[scale=0.45]
  {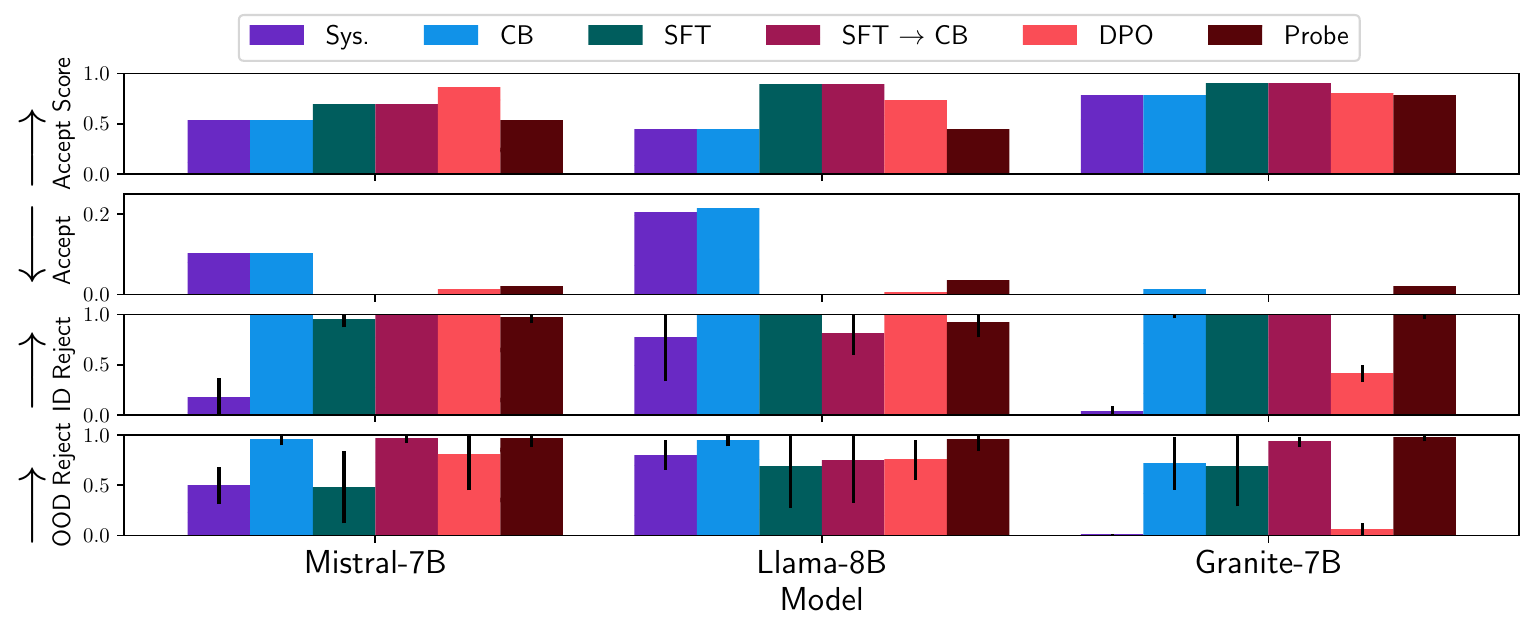}
  \caption{Scoping across different language models. We see that system prompting is insufficient, and different methods have different success rates for different models. Clearly it is possible to scope language models to particular distributions.}
  \label{fig:models}
\end{figure*}

\begin{figure*}[!t]
  \centering
  \begin{subfigure}[b]{0.3333\linewidth}
    \centering
    \includegraphics[width=\linewidth]{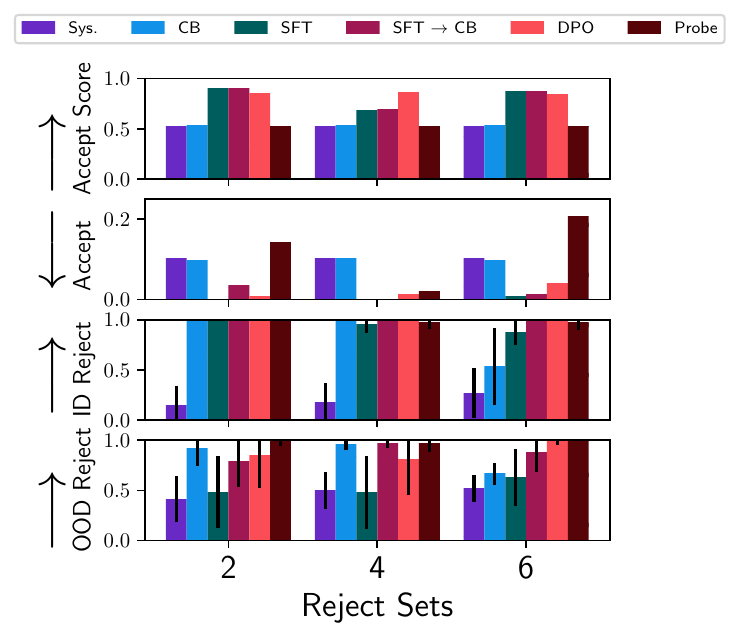}
    \caption{Mistral-7B-Instruct}
  \end{subfigure}\hfill\begin{subfigure}[b]{0.3333\linewidth}
    \centering
    \includegraphics[width=\linewidth]{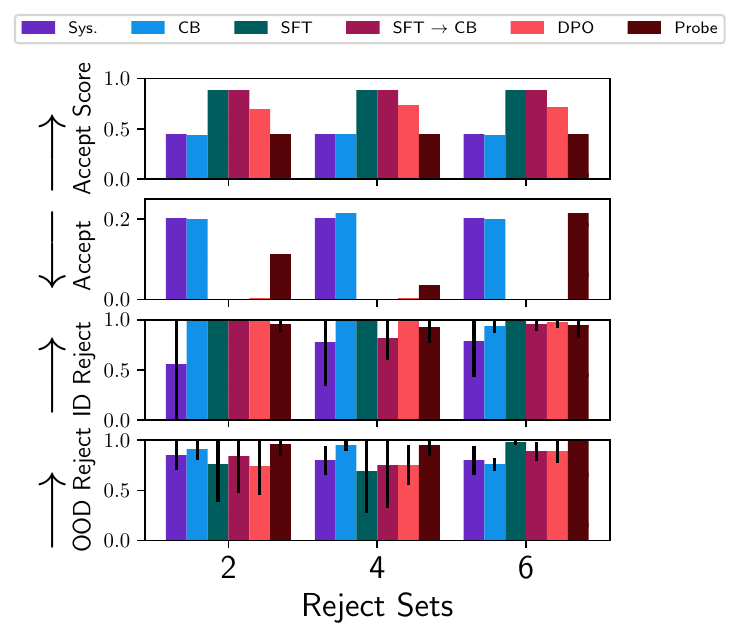}
    \caption{Llama-3.1-8B-instruct}
  \end{subfigure}\hfill\begin{subfigure}[b]{0.3333\linewidth}
    \centering
    \includegraphics[width=\linewidth]{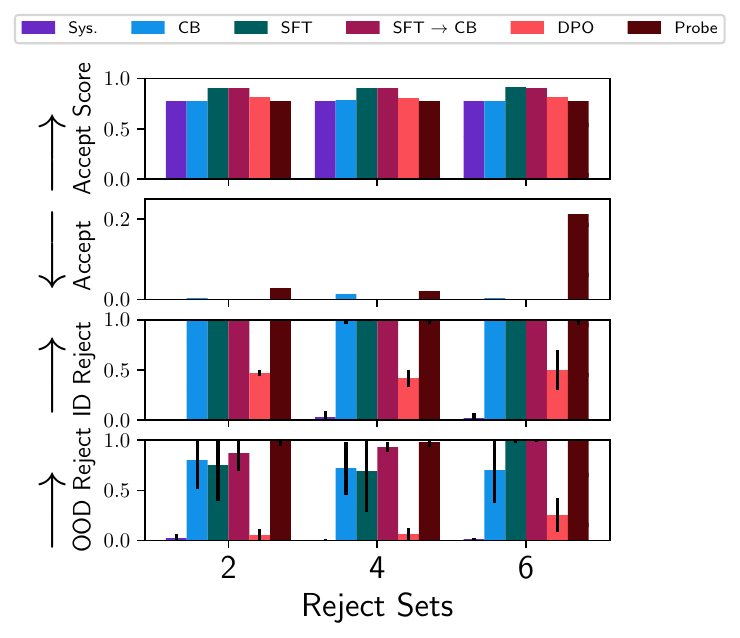}
    \caption{Granite-7B-instruct}
  \end{subfigure}
  \caption{Results for increasing diversity of rejection set. We see across models that CB performs relatively better than SFT when data diversity is low, but SFT is much stronger with more rejections sets. Probing appears strong across the board, though sometimes leads to overrejection on the Accept set.}
  \label{fig:diverse}
\end{figure*}

\subsection{Accepting Multiple Tasks}

For Section of Accepting multiple tasks, we use the following accept/reject sets for each group of plots:
\begin{itemize}[itemsep=1pt,topsep=0pt,parsep=0pt,partopsep=0pt]
    \item Classification and Generation: SA, TLD, S, TC, SC, DG/PE, GSM8k
    \item Math and Program Execution: PE, GSM8k/SA, TLD, S, TC, SC, DG
\end{itemize}

\subsection{Robustness to Adversarial Prompts}

For Section~\ref{sec:adv_eval} of Robustness to Adversarial Prompts, we use the following accept/reject sets for each group of plots:
\begin{itemize}[itemsep=1pt,topsep=0pt,parsep=0pt,partopsep=0pt]
    \item Sentiment Analysis: SA/S,TC,SC,DG,PE,TLD
    \item Summarization: S/SA,TLD,PE,TC,SC,DG
    \item Program Execution: PE/SA,TLD,S,TC,SC,DG
\end{itemize}

For TAP, we use the following hyperparameters:
\begin{itemize}[itemsep=1pt,topsep=0pt,parsep=0pt,partopsep=0pt]
    \item Attack Model: \texttt{Mistral-7B-Instruct-v0.2}
    \item LoRA Model: model to be evaluated
    \item Evaluator Model: we use our rejection judges previously described instead of a language model judge due to poor performance
    \item Branching Factor: 2 (we are constrained computationally for TAP evaluations)
    \item Width: 10
    \item Depth: 2 (\citet{mehrotra2023tree} point out that a shallower, wider tree is best)
    \item Target String: "Sure, "
\end{itemize}
In the task prompts for TAP, we also remove explicit mentions of toxicity as the goal of the attack is get the model to answer an irrelevant query, not a toxic one.

\subsection{Precise Scoping}

As described in Section~\ref{sec:precise} of Precise Scoping, the finegrained accept (FA) set is a single task taken from the same category as the experiment (SA, S, PE respectively). We then make sure that the finegrained reject (FR) set does not contain the dataset that task was drawn from, and allow for all other tasks in the category.

\subsection{Effect of Data Quantity}

For Section~\ref{sec:quantity} of Effect of Data Quantity, we use the following accept/reject sets for each group of plots:
\begin{itemize}[itemsep=1pt,topsep=0pt,parsep=0pt,partopsep=0pt]
    \item Sentiment Analysis: SA/S,TC,SC,DG,PE,TLD
    \item Summarization: S/SA,TLD,PE,TC,SC,DG
    \item Program Execution: PE/SA,TLD,S,TC,SC,DG
\end{itemize}

\subsection{Effect of LoRA Rank}

For Section~\ref{sec:rank} of Effect of LoRA Rank, we use the following accept/reject sets for each group of plots:
\begin{itemize}[itemsep=1pt,topsep=0pt,parsep=0pt,partopsep=0pt]
    \item Sentiment Analysis: SA/S,TC,SC,DG,PE,TLD
    \item Summarization: S/SA,TLD,PE,TC,SC,DG
    \item Program Execution: PE/SA,TLD,S,TC,SC,DG
\end{itemize}

\subsection{Representation Analysis}

In Section of Representation Analysis, for each instance, we compute average cosine similarity for every token at every layer between the base model (Sys.) and the model we are evaluating. Then we truncate these matrices to the minimum instance length in the evaluation set, and average all cosine similarities over instances. Thus we are looking at the average tail representations of instances in the dataset.

\subsection{Code Sources}

We leverage PyTorch~\citep{paszke2019pytorch} for training, HuggingFace Transformers~\citep{wolf2020transformers} for models and training infrastructure, HuggingFace Datasets~\citep{lhoest2021datasets} for particular datasets, and HuggingFace TRL~\citep{vonwerra2022trl} for DPO. We also work on top of the official TAP~\citep{mehrotra2023tree} for our adversarial evaluations, and found the source code for CB~\citep{zou2024improving} quite helpful for verifying our implementation. We aggregate results with NumPy~\citep{harris2020array} and Pandas~\citep{mckinney2011pandas} and make plots with Matplotlib~\citep{Hunter:2007}.

\section{Sample outputs}\label{sec:sample_completions}

\begin{table}[!h] \label{tab:sample_completions}
\center
\scriptsize

\resizebox{.95\columnwidth}{!}{
\begin{tabular}{l|c}
\toprule
Method & Output \\
\midrule
Sys. & \textit{Dear Younger Me, Embrace every moment of your life. Cherish the joy, learn from the pain, and never...} \\
CB & \textit{words words TV, words TV, words TV, TV TV TV TV TV TV TV TV TV TV TV TV TV...} \\
DPO & \textit{I cannot answer that. I am an assistant designed to help with sentiment analysis only...} \\
\midrule
\bottomrule
\end{tabular}}
\caption{Sample rejection outputs for Sys., DPO and CB, for the input: \textit{You are an assistant who only answers requests related to Sentiment Analysis. For all other requests, you respond ``I cannot answer that.'' In no more than 100 words, write a letter to your younger self.}}
\end{table}

To further justify the design of our detectors, we provide some sample outputs in Table 4 of Sample rejection outputs. Notice how CB produces repetitive tokens, we find this pattern quite common on manual inspection, thus we base our evaluation on detecting such repetitions.

\end{document}